# Autonomous Road Vehicle Emergency Obstacle Avoidance Maneuver Framework at Highway Speeds

Evan Lowe Levent Güvenç

(*lowe.500@osu.edu*) (*guvenc.1@osu.edu*)

The Ohio State University, Automated Driving Lab

## Abstract

An Autonomous Road Vehicle (ARV) can navigate various types of road networks using inputs such as throttle (acceleration), braking (deceleration), and steering (change of lateral direction). In most ARV driving scenarios that involve normal vehicle traffic and encounters with vulnerable road users (VRUs), ARVs are not required to take evasive action. This paper presents a novel Emergency Obstacle Avoidance Maneuver (EOAM) methodology for ARVs traveling at higher speeds and lower road surface friction, involving time-critical maneuver determination and control. The proposed EOAM Framework offers usage of the ARV's sensing, perception, control, and actuation system abilities as one cohesive system, to accomplish avoidance of an on-road obstacle, based first on performance feasibility and second on passenger comfort, and is designed to be well-integrated within an ARV high-level system. Co-simulation including the ARV EOAM logic in Simulink and a vehicle model in CarSim is conducted with speeds ranging from 55 – 165 km/h and on road surfaces with friction ranging from 1.0 to 0.1. The results are analyzed and given in the context of an entire ARV system, with implications for future work.

# 1. Introduction

Though automotive safety technology has improved substantially over recent decades, there are still several accidents each year, with fatalities of approximately 1.25 million people annually (World Health Organization, 2018) (Centers for Disease Control and Prevention (CDC), 2020). The National Motor Vehicle Crash Causation Survey, conducted from 2005 to 2007, showed that 94% of all vehicle crashes in the US were due to driver error (Singh, 2015). Of those driver-related errors, the largest percentage (41%) of the human error was due to Recognition Errors, which can include driver inattention, internal and external distractions, and inadequate surveillance. This was followed by Decision Errors (33%) such as driving too fast for the existing road conditions, and misjudgment of either the gap between vehicles or other vehicle speeds. These sobering statistics demonstrate that human driving error causes many car accidents, which are often fatal, even with the latest vehicle safety improvements. These statistics grow in severity as a) vehicle speeds increase, and b) road and/or environmental conditions degrade.

One predominant takeaway from these collective statistics is that an Autonomous Road Vehicle (ARV) could be well-suited to safely deal with these driving situations that require evasive maneuvering, with high precision and accuracy, full utilization of the associated vehicle dynamics, and comprehensive real-time data regarding the outside environment. In this context, an ARV is equipped with Level 3-5 autonomy, as defined by the Society of Automotive Engineers J3016 standard (SAE, 2016). An ARV with an exclusive subsystem designed to handle Emergency

Obstacle Avoidance Maneuvers (EOAMs) would be ideal to specialize in more severe ARV obstacle avoidance tasks.

Regarding obstacle avoidance maneuvers for ARVs, there are several works within the past 30 years that detail useful inputs for an autonomous vehicle to complete a lane change during severe lateral dynamic conditions. An earlier emergency maneuver study by (Chee & Tomizuka, 1994) tested various types of steering controllers that utilized input data that could be extracted from onboard vehicle sensors without asserting decision-making logic for the ARV. Later updates provided by (Shiller & Sundar, 1998) and (Hattori, Ono, & Hosoe, 2006) included specific decision-making logic based on vehicle states relative to sensed road objects, such that the autonomous vehicle may either brake, steer and brake, or provide a pure steering maneuver, to avoid the road-going object ahead.

In 2020 (Peng, et al., 2020) offered a scoring-based system that utilized a minimum safety distance, and required steering frequency with maximum lateral acceleration, to assess emergency maneuver safety and comfort score. Additionally, (Zhu, Gelbal, Aksun-Güvenc, & Güvenc, 2019) provided a lookup table approach to create an obstacle avoidance trajectory for an ARV with parameter-space robustness, that could be completed in real-time, and this approach was implemented using rapid prototype hardware in a HIL environment, then on an actual vehicle. Ding et al used a combined geometric and kinematic approach to trat the same problem (Ding, et al., 2020).

It is noted that many different approaches can work for an ARV in an emergency maneuver situation. What is novel in this paper is the implementation of a particular decision-making and trajectory-generation approach within the context of an entire ARV high-level system, which can work faster than real-time and can, therefore, easily be integrated into a production-level autonomous road vehicle (ARV) system framework. This paper preserves recognition of an ARV system hierarchy when determining the time and conditions in which the emergency maneuver should occur, the control handoff from the high-level ARV system to the specific ARV emergency maneuver domain, and the return of vehicle control back to the high-level ARV system, after the maneuver is complete. This end goal of a production-intent ARV emergency maneuver framework is unique in its active integration within a high-level ARV system architecture.

The organization of the rest of the paper is as follows. Sections 2 – 4 introduce the ARV EOAM Framework design goals, system description, and modeling details. EOAM Framework decision-making logic creation details are presented in Sections 5 – 7, along with more additional EOAM Framework novelties presented in Section 8. The simulation scenario premise and results are presented in Sections 10 – 11, while the final conclusions and recommendations for future work are provided in Section 12.

## 2. Overall EOAM Framework Goals

A general ARV EOAM framework system should utilize the key ARV characteristics shown in Figure 1 parsing of relevant World Model Data, Decision-Making Subsystem, Trajectory Generation Subsystem, Trajectory-Tracking Control Subsystem, and Actuation (throttle, steering, and brakes).

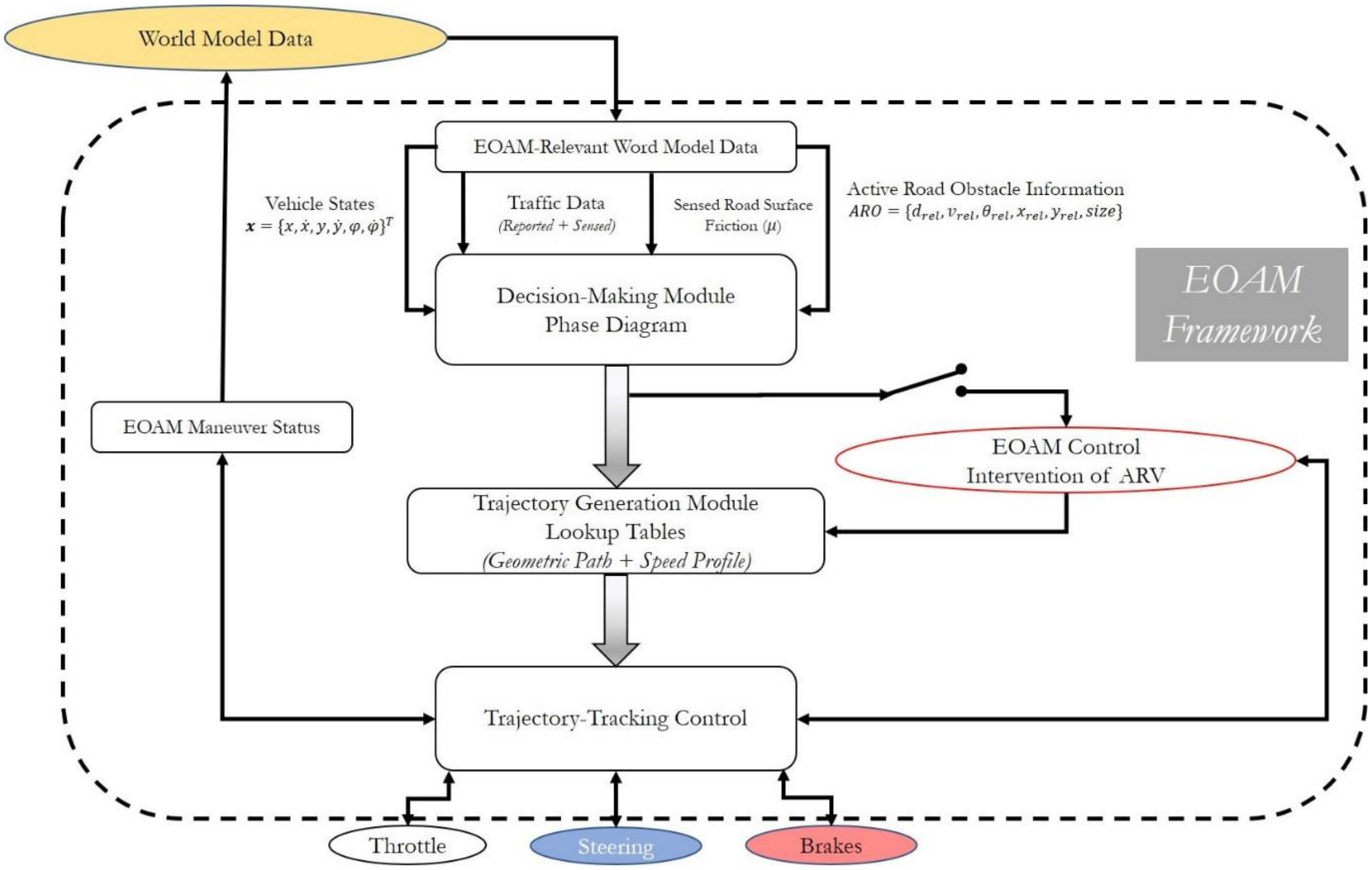

*Figure 1: Summary EOAM Framework system diagram*

This EOAM framework should also embrace operational design domain (ODD) and dynamic driving task (DDT)-based architecture (Serban, Poll, & Visser, 2018), (Mody, et al., 2018), (Gyllenhammar, et al., 2020) (Griffor, Wollman, & Greer, 2021) that allows for specific domains to take care of unique tasks that occur less frequently than normal driving conditions; one of the domains could manage emergency maneuvers involving large magnitude lateral dynamics with low processing and execution times.

The EOAM Framework described in this paper is designed to be a real-time capable, domain-based system, which can perform successful maneuvers during ARV travel at highway speeds (55 km/h to at least 120 km/h) (Euro NCAP, 2020) on surfaces that range from ideal (dry, $\mu = 1.0$), to slippery (ice, $\mu = 0.1$). For an ARV to execute an EOAM, the primary ARV systems must all function in harmony with the EOAM domain; these systems are Sensing, Perception, Decision-Making, Control, and Actuation.

The overall goals for this EOAM Framework are as follows:

- Understand and recognize World Model conditions that require an EOAM; this includes the ego ARV states and ARO (active road object) relative states.
- Create and utilize decision-making logic that allows the ARV to know what type of maneuver to perform, and precisely when to perform it, based on the current environmental conditions, known vehicle traits, current on-road activity including the states of the object(s) that must be avoided (the ARO).
- Through ARV actuator (steering, throttle, brake) control, perform the safest emergency maneuver based on environmental, ARV, and ARO states.
- Provide warnings to the vehicle occupants based on ARV and ARO states, and potentially before the EOAM begins, and which can also activate pre-collision safety measures, such as occupant safety belt pre-tensioners, airbag arming, and increased seat side and leg bolstering (Pack, Koopmann, Yu, & Najm, 2005) (Cho, Choi, Shin, & Yun, 2010).

# 3. EOAM Framework System Description

## 3.1 EOAM Framework: Logic Summary

The software logic for this EOAM Framework was deployed in Simulink with some Matlab functions, then co-simulated with CarSim version 2021.0. This simulation ran faster than in real-time as a confirmation that this same software logic could be deployed in actual vehicle hardware within an ARV, for real-time decision-making and maneuver execution. Below is a summary of the steps utilized in the EOAM Framework logic used in this research, based on the system information in Figure 1:

1. Determine if current conditions are necessary for an EOAM, using the decision-making-module (DMM) phase.
2. When a lane change is required, utilize 3D lookup tables that were computed offline, to determine the vehicle's desired longitudinal acceleration and steering input through the maneuver.
3. Utilize steering and longitudinal acceleration controllers to assure the maneuver is robust enough to complete the maneuver as prescribed.
4. Monitor the ARV's lateral position concerning a point-of-no-return, after which the vehicle will attempt completion of the maneuver, even if oncoming traffic is detected.
5. Provide a specified time-based duration for the entire lane change maneuver to complete that allows the ARV to reach a stable final position, void of transient lateral motion.
6. After the maneuver is completed, return to the original lane of travel and hand control back to the ARV high-level system controller.

### 3.2 EOAM Framework: Discrete Modes-of-Operation

Overall, there are six discrete modes-of-operation that are conducted by the EOAM Module within the ARV system architecture:

0) NORMAL: the high-level ARV system is in control as traffic conditions do not require EOAM intervention.
1) UPDATE, BRAKE: EOAM intervention with occupant warning is required and the ARV-ARO location in the DMM phase diagram is within a pure braking section, no oncoming vehicle is detected.

2) UPDATE, STEER + BRAKE: an EOAM lane change maneuver with occupant warning, known geometric path, and longitudinal acceleration profile is required based on the ARV-ARO location in the DMM phase diagram, no oncoming vehicle is detected.
3) ONCOMING, BRAKE: an oncoming vehicle is detected at the time of either 1) or 2), but before the ARV reaches the lateral point-of-no-return, so limit braking should be conducted.
4) ONCOMING, STEER + BRAKE: an oncoming vehicle is detected at the time of either 1) or 2), and after the ARV reaches the lateral point-of-no-return, so the lane change maneuver should continue so that the ARV avoids an offset/oblique collision with the first ARO; once the ARV reaches the desired position in the next lane, it should immediately return to the original lane, to ideally avoid the detected oncoming traffic. After the ARV returns to the desired lane position in the original lane, it should hand control back over to the high-level ARV system
5) RETURN: The ARV has completed 2) or the ARV has reached the desired lane position after 4) occurred, so, it should return to the original lane. After the ARV returns to the desired lane position in the original lane, it should hand control back over to the high-level ARV system.

A summary logic flowchart for the EOAM Framework proposed in this paper that combines the high-level EOAM system summary and EOAM Framework discrete modes of operation is shown in

Figure 2.

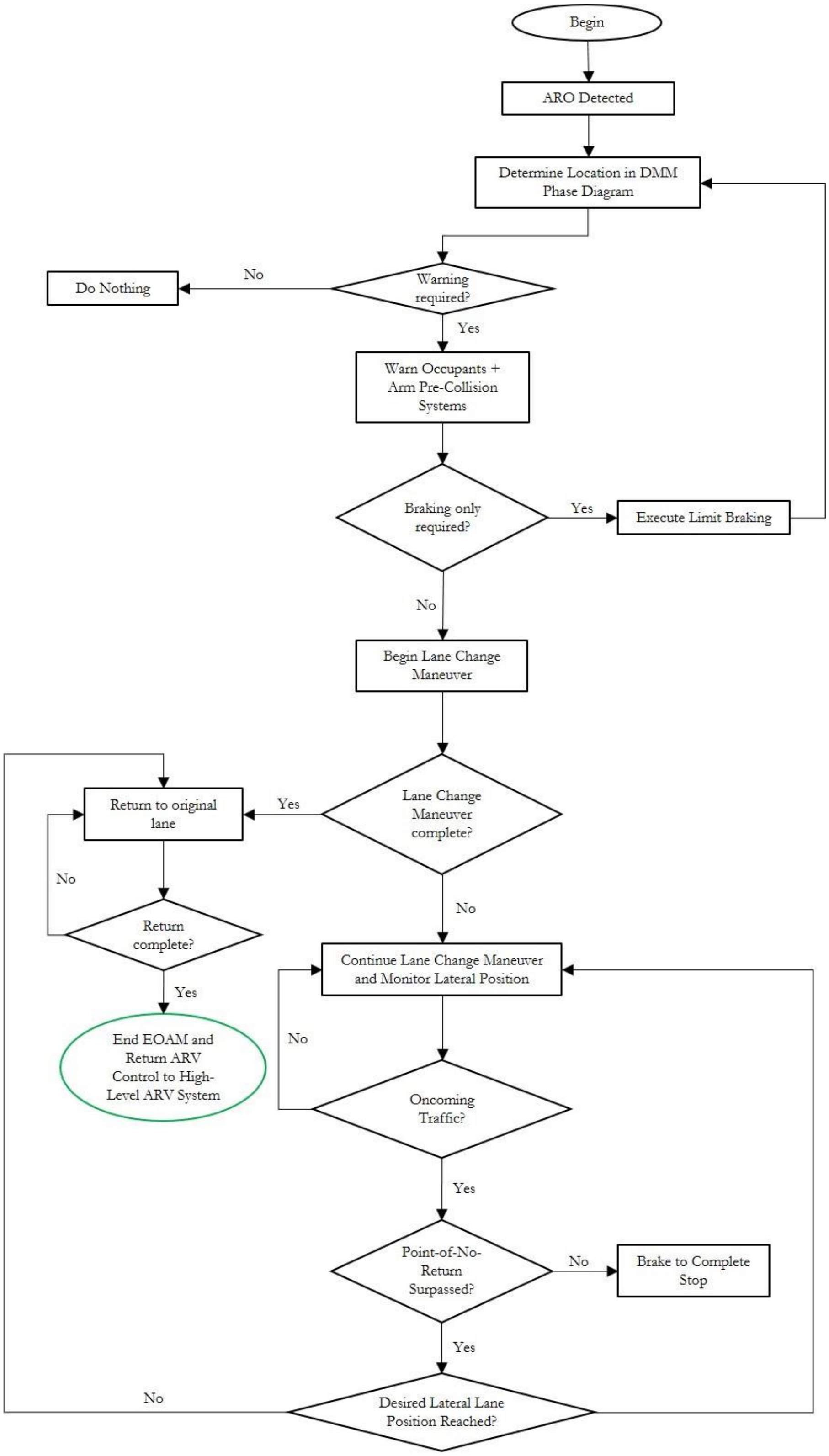

*Figure 2: EOAM logic flowchart*

# 4. Modeling Details

## 4.1 Vehicle Modeling

For simulation of an EOAM under varying surface friction (mu) conditions at highway speeds (55 to 165 km/h), a simplified lateral dynamic model is useful in that it can be linearized about a single longitudinal speed and can capture transient lateral motion, with relatively low computational complexity. The vehicle used in the numerical work in this paper is an E-Class sedan with front-wheel steering and front-engine/rear-wheel-drive powertrain layout, without an anti-lock braking system (ABS) or electronic stability control (ESC). The equations of motion for the planar bicycle model concerning the vehicle body can be derived by first creating a free body diagram (as shown in

Figure 3) that includes the relevant forces and moments on the vehicle during a turning maneuver, such that Newton’s second law applies (Wong, 2008). The notation used in

Figure 3 is given in Table 1.

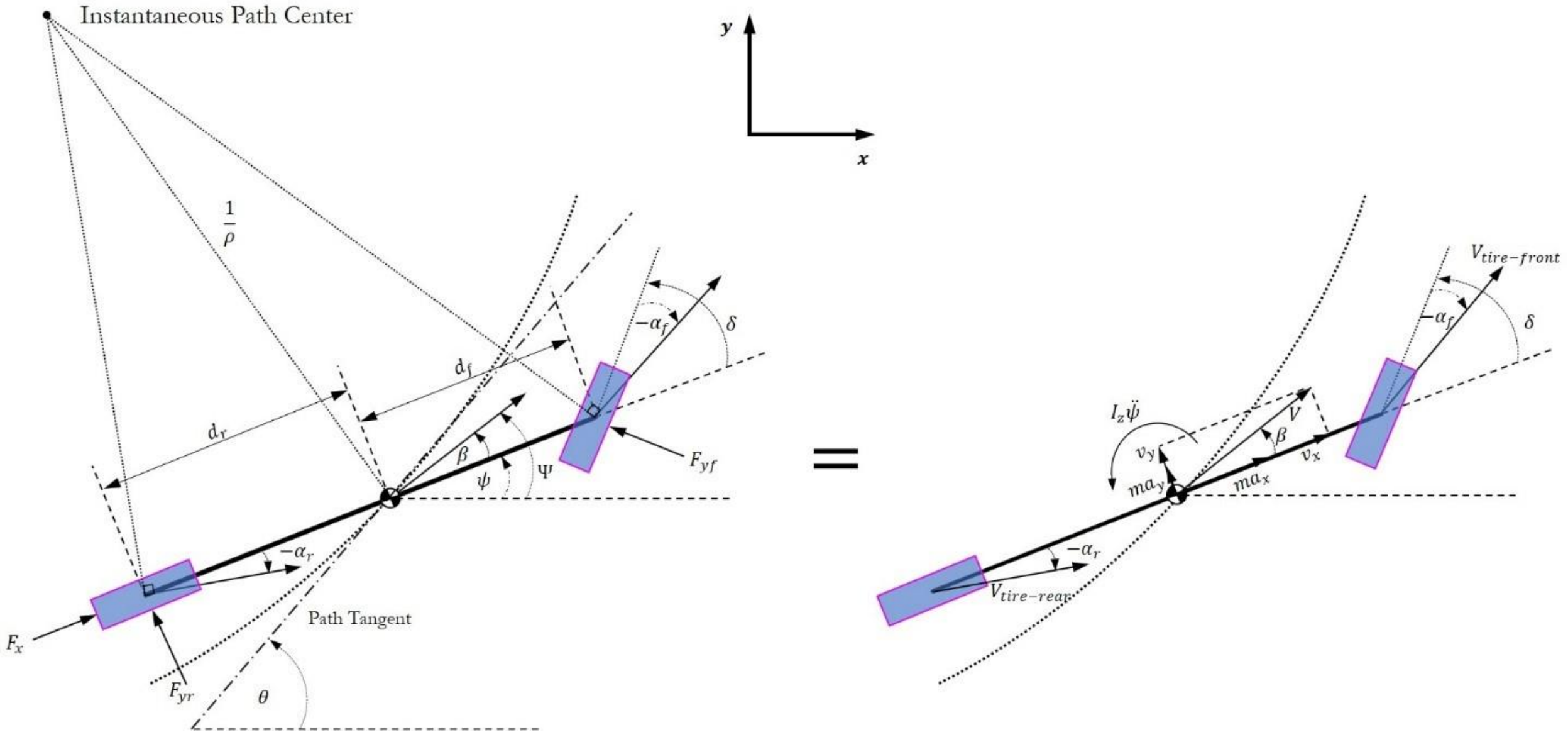

*Figure 3: Free body diagram representing Newton's second law with respect to a 3 DOF bicycle model*

*Table 1: Relevant nomenclature for 3 DOF planar vehicle model used in this paper*

| | |
|---|---|
| $F_x$ = longitudinal (tractive) force | $\beta$ = body (side)slip angle |
| $F_{yf}$ = lateral force on the front axle | $\alpha_f$ = front slip angle |
| $F_{yr}$ = lateral force on the rear axle | $\alpha_r$ = rear slip angle |
| $a_x$ = longitudinal acceleration | $\psi$ = body-fixed yaw angle |
| $a_y$ = lateral acceleration | $\dot{\psi}$ = body-fixed yaw rate |
| $d_f$= distance from the front axle to the center of gravity | $\ddot{\psi}$ = body angular acceleration |
| $d_r$= distance from the rear axle to the center of gravity | $\delta$ = steering road wheel angle |
| $m$ = mass | $\Psi$ = absolute (global) ARV heading angle |
| $I_z$ = mass moment of inertia about vehicle z-axis | $\theta$ = absolute (global) path yaw angle |

The forces can be summed in the longitudinal and lateral directions with respect to the axes fixed to the vehicle body, in addition to the moments summed about the vehicle's center of gravity (Wong, 2008):

$$F_x - F_{yf} sin\ (\delta) = ma_x \tag{1}$$

$$F_{yr} + F_{yf} cos\ (\delta) = ma_y \tag{2}$$

$$d_{\mathrm{f}} F_{yf} cos\ (\delta) - d_r F_{yr} = I_z \ddot{\psi} \tag{3}$$

where the longitudinal and lateral acceleration vectors contain translational and rotational components

$$a_x = \dot{v}_x - v_y \dot{\psi} \tag{4}$$

$$a_y = \dot{v}_y + v_x\dot{\psi} \tag{5}$$

The cornering stiffness for the front and rear tires were extracted by reviewing the linear slope of the lateral force versus slip curves (Gillespie, 1992) from the CarSim 225/16 R18 Touring Tire data (Figure 4).

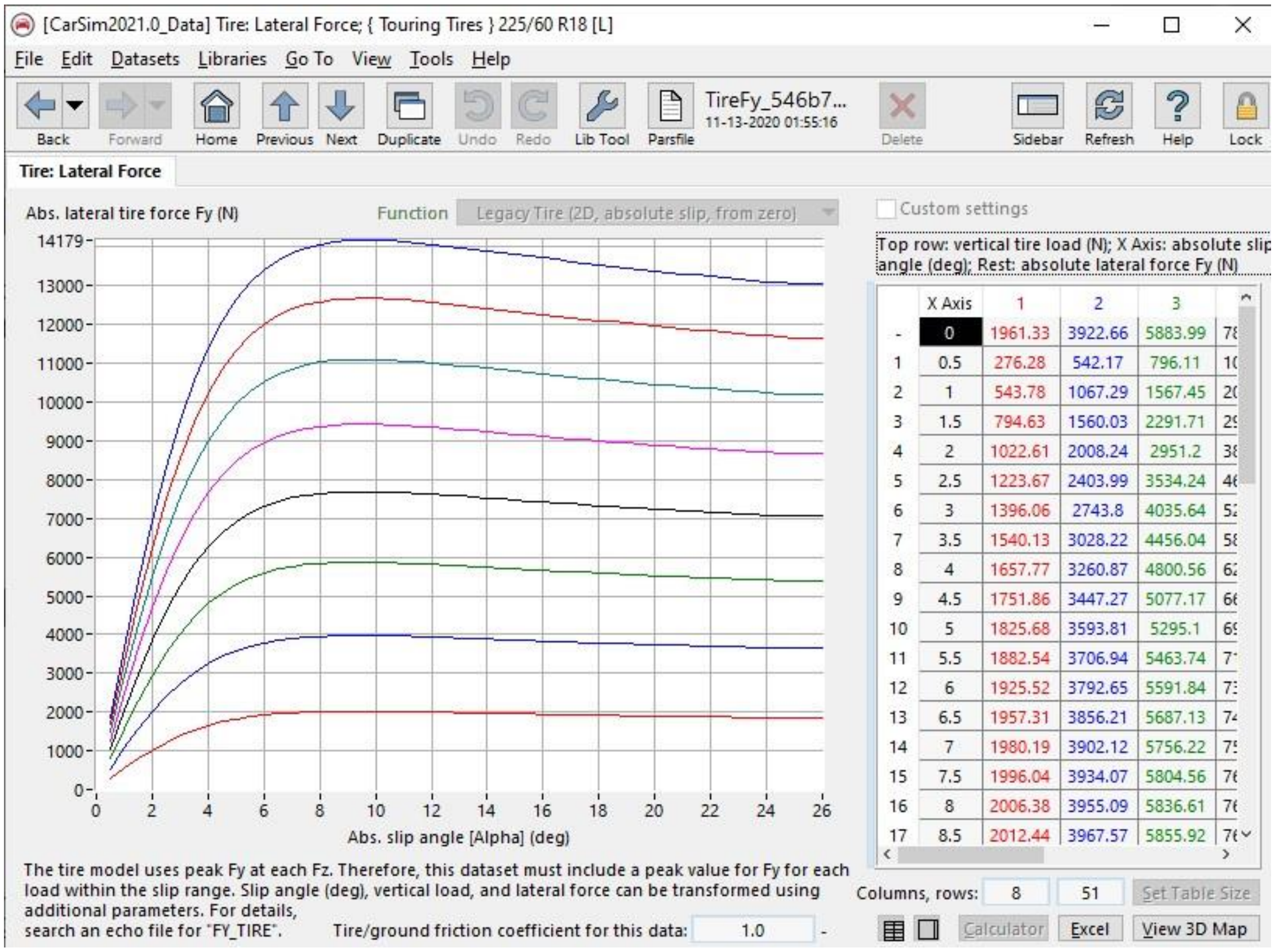

*Figure 4: Tire lateral force versus slip angle curves for simulated E-class sedan, (Mechanical Simulation, 2020)*

The maximum lateral tire forces were approximated as linear using linear lateral tire cornering stiffness ($C_\alpha$) and lateral tire slip angle ($\alpha$), using common formulations in this area (Gillespie, 1992), (Milliken & Milliken, 1995), (Wong, 2008), (Brach & Brach, 2009), (Brach & Brach, 2011). This linear approximation is represented by the following equation:

$$F_{y\,max} = -C_\alpha \alpha^* \qquad (6)$$

In the trajectory creation section below, this approximation is used as the lateral force output whenever the actual tire slip angle is greater than or equal to the maximum front or rear tire lateral slip angle of 5 degrees ($\alpha_{max} = \alpha^* = 5°$). Further details regarding this modeling can be found in (Lowe, 2022).

### 4.2 Lane Change Geometric Trajectory Creation

A fifth-order polynomial geometric trajectory and constant speed profile comprised the initial trajectory, noting that a coupled path and speed or acceleration profile equates to a trajectory. A fifth-order polynomial was implemented in the EOAM methodology as it is geometrically useful for a single lane change (SLC) maneuver, has geometric G3 continuity, and has a continuous curvature rate when parameterized in terms of arc length.

The initial formulation for this fifth-order polynomial trajectory includes a time-based output (C2 parametric continuity) though this path is later parameterized in terms of arc length, so it is ultimately considered with geometric (G3) continuity. Details of how a fifth-order polynomial can be used as a formulation for an ARV path generation are noted in (You, et al., 2015) and (Mehmood, Liaquat, Bhatti, & Rasool, 2019). If, for example, the desired trajectory had a constant longitudinal speed of 20 m/s, maneuver time of ($t_f$) 2.5 s, beginning with zero lateral motion, and desired final y-position ($y_f$) of 3.5 m, with zero lateral motion in its final conditions, the output would result in those shown in

Figure 5.

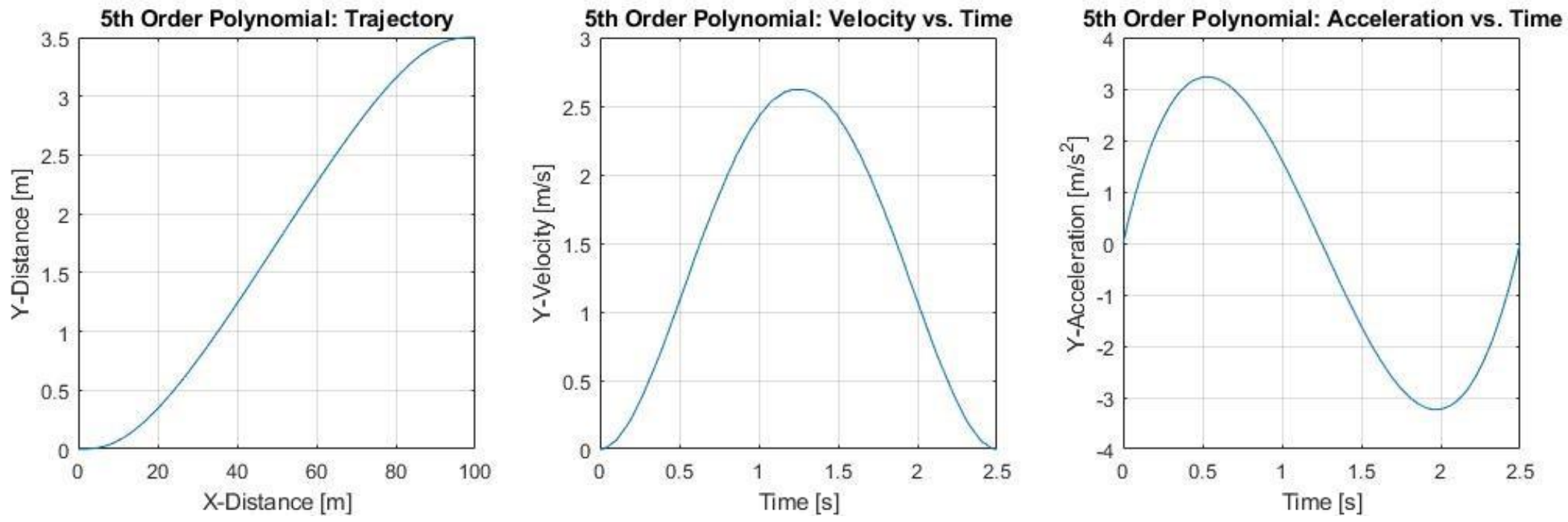

*Figure 5: Example fifth-order polynomial output trajectory and corresponding time derivatives*

From here, the trajectory can be parameterized in terms of arc length, which will be explained in the next subsection.

### 4.3 Trajectory Parameterization

To prepare the determined trajectory for usage in inverse dynamics calculations to output steering road wheel angle and subsequent longitudinal acceleration profile optimization, the recently found time-based trajectory should be parameterized in terms of arc length, $s$. This can be done with the following formulation:

$$\vec{s}(t) = \begin{bmatrix} \vec{x}(t) \\ \vec{y}(t) \end{bmatrix} \tag{7}$$

$$\|\vec{s}(t)\| = \int_{t_0}^{t_f} \sqrt{\left(\frac{dx}{dt}\right)^2 + \left(\frac{dy}{dt}\right)^2}\, dt \tag{8}$$

where $s$ is the time integral of the two-norm of the time derivatives of $x(t)$ and $y(t)$. Also useful for input into the inverse dynamics calculations is the path curvature:

$$K(s) = \frac{\vec{s}' \text{ x } \vec{s}''}{\|\vec{s}'\|} \left| \frac{y''(x)}{(1 + (y'(x))^2)^{3/2}} \right| = \left| \frac{\frac{dx}{dt} * \frac{d^2y}{dt^2} - \frac{dy}{dt} * \frac{d^2x}{dt^2}}{\left( \left(\frac{dx}{dt}\right)^2 + \left(\frac{dy}{dt}\right)^2 \right)^{3/2}} \right| = \frac{d\theta}{ds} = \theta'(s) \qquad (9)$$

When considering the angle which the trajectory makes with the absolute axes in terms of arc length, $\theta(s)$, the following formulations of $x(s)$ and $y(s)$ can be determined:

$$\theta(s) = \int_0^s K(\tau) d\tau \qquad (10)$$

$$x(s) = x_0 + \int_0^s cos\ (\theta(\tau)) d\tau \qquad (11)$$

$$y(s) = y_0 + \int_0^s sin\ (\theta(\tau)) d\tau \qquad (12)$$

It should be noted that

$$K(s) = \frac{1}{R} = \rho \qquad (13)$$

where $R$ is the path radius of curvature.

## 4.4 Optimization Routine Summary Outputs

The inverse dynamics premise used in this research is similar to that outlined by (Shiller & Sundar, 1995) (Lowe, 2022), and those formulation details will not be covered here for the sake of brevity. The inverse dynamics solution outputs for the front and rear slip angles, front and rear lateral tire forces, and steering road wheel angle versus arc length when the initial ARV-ARO relative speed is 100 km/h is illustrated in

Figure 6.

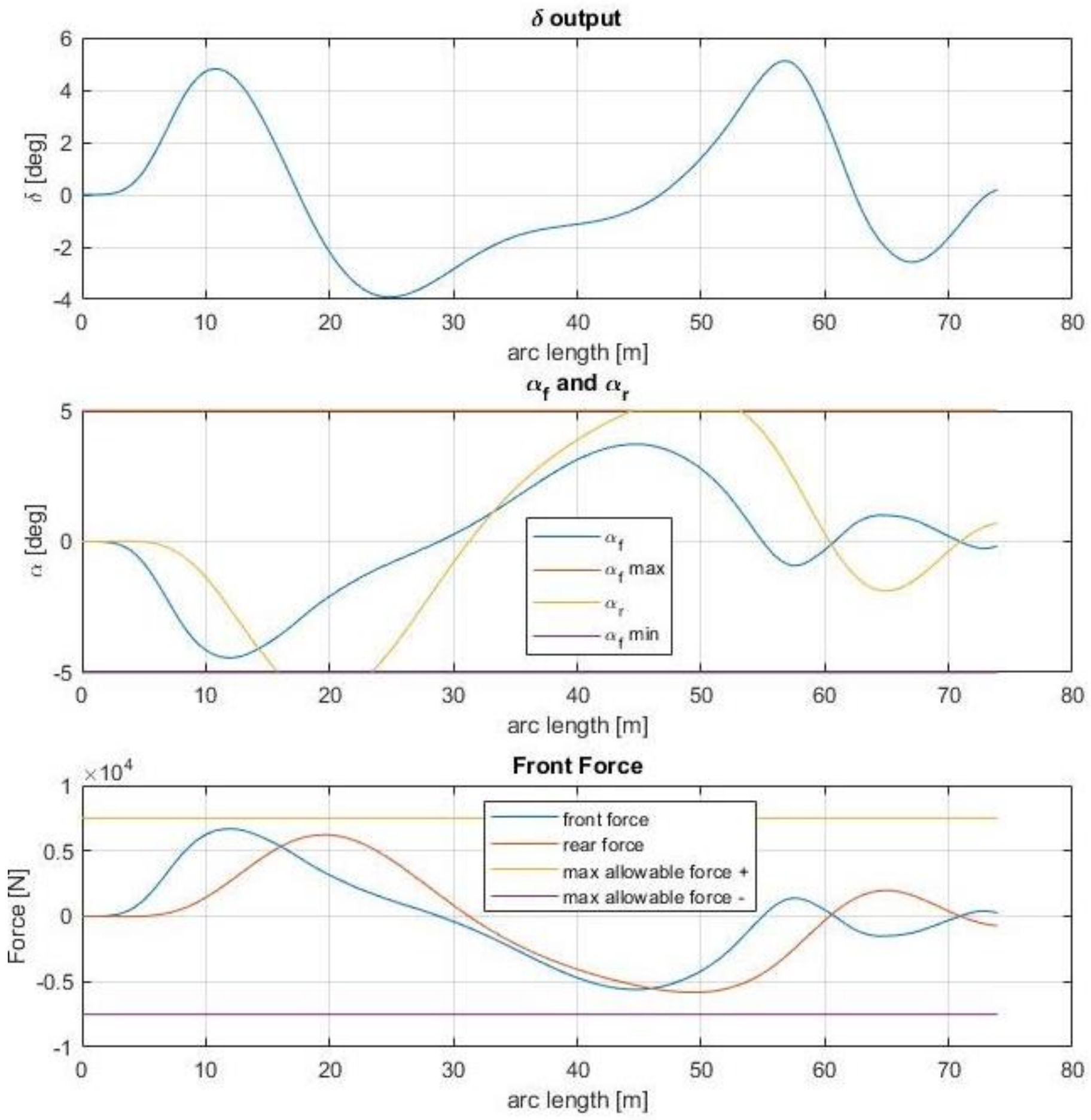

*Figure 6: Steering road wheel angle, front and rear slip angles, and front and rear force outputs from inverse dynamics calculations*

The final front and rear lateral forces and slip angles can then be used as inputs to determine the longitudinal acceleration constraints, as shown in

Figure 7 (with the minimum and maximum longitudinal acceleration constraints along with a zero-acceleration example profile for reference).

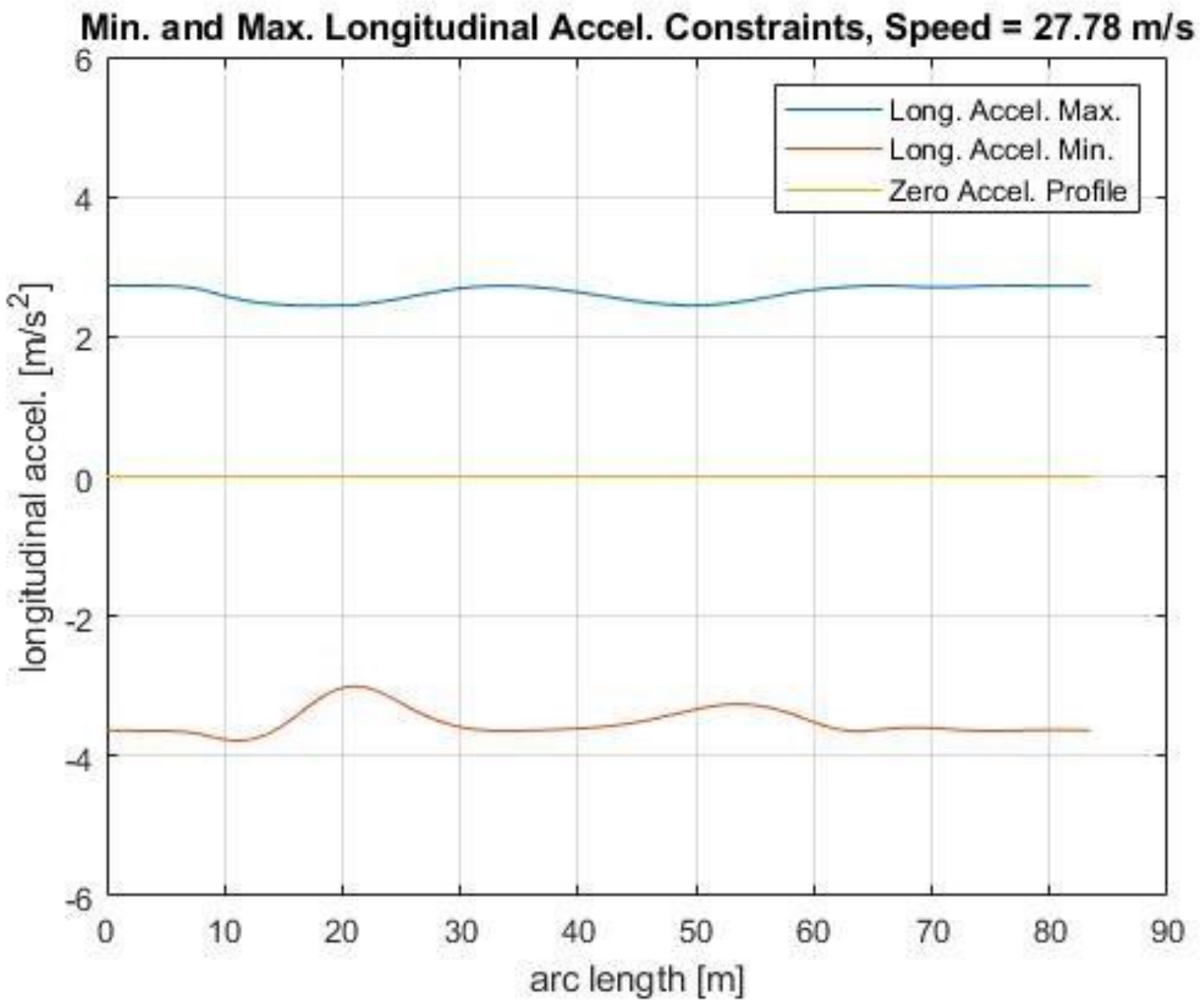

*Figure 7: Minimum and maximum longitudinal acceleration constraints based on lateral slip angles and forces generated by the steering input*

Once the input steering road wheel angle and the longitudinal acceleration constraints are determined, an optimal or suboptimal longitudinal acceleration profile can be calculated using time-based optimization. This should minimize the total longitudinal distance needed to complete the lane change maneuver. This optimization includes the lateral dynamic constraints of the vehicle based on its relevant parameters (mass, yaw moment of inertia, longitudinal center of gravity location, front and rear tire cornering stiffness, etc.) as well as the lateral and longitudinal force constraints of the tires.

## 4.5 Optimization for Longitudinal Acceleration Profile Creation: Overview

To reiterate, the goal of the nonlinear constrained optimization is to minimize the longitudinal distance traveled by the ARV during the entire EOAM, based on the input steering angle and vehicle dynamic constraints, and control input constraints. Thus, the cost function for this optimization includes the longitudinal distance from the initial time of the maneuver to the final time at which the maneuver is complete. The output of the optimization is the longitudinal acceleration profile that represents the minimum distance traveled by the vehicle, during the EOAM, given the constraints, boundary conditions, and vehicle states.

When considering this EOAM distance minimization optimal control problem, the ARV states are the longitudinal and lateral distances and speeds, yaw, and yaw rate, $\boldsymbol{x} = \{x, \dot{x}, y, \dot{y}, \psi, \dot{\psi}\}$, and inputs of tractive force and steering road wheel angle, $\boldsymbol{x} = \{F_t, \delta\}$. The optimization formulation is as follows:

Cost function:

$$\min_u J = x_1(t_f) = \int_0^{t_f} x_2(\boldsymbol{x}, \boldsymbol{u}, t)dt \quad (14)$$

with initial ARV-ARO relative speed at the beginning of the EOAM, $\dot{x}_0$, and free final time, $t_f$, final yaw, $\psi(t_f)$, and final yaw rate, $\dot{\psi}(t_f)$, subject to the system dynamics

$$\dot{x}_1 = x_2 \quad (15)$$

$$\dot{x}_2 = \frac{1}{m}[\cos(x_5) - F_r \sin(x_5) - F_f \sin(x_5 + u_2)] \quad (16)$$

$$\dot{x}_3 = x_4 \quad (17)$$

$$\dot{x}_4 = \frac{1}{m}[F_r \cos(x_5) + u_1 \sin(x_5) - F_f \cos(x_5 + u_2)] \quad (18)$$

$$\dot{x}_5 = x_6 \quad (19)$$

$$\dot{x}_6 = \frac{1}{I_z}[-d_r F_r + d_f F_f \cos(u_2)] \quad (20)$$

boundary conditions

$$x_1(0) = 0 \tag{21}$$

$$x_2(0) = \dot{x}_0 \tag{22}$$

$$x_3(t_f) = y_f \tag{23}$$

$$x_4(0) = x_5(0) = x_6(0) = 0 \tag{24}$$

state constraints

$$h(\boldsymbol{x}) = F_f(\boldsymbol{x}) - F_{f\,max}(\boldsymbol{x}) \leq 0 \tag{25}$$

and control constraints

$$g_1(\boldsymbol{u}) = u_2 - \delta_{max} \leq 0 \tag{26}$$

$$g_2(\boldsymbol{u}) = \delta_{min} - u_2 \leq 0 \tag{27}$$

$$g_3(\boldsymbol{x},\boldsymbol{u}) = \left(\frac{u_1}{F_{x\,max}}\right)^2 + \left(\frac{F_r(\boldsymbol{x})}{F_{y\,max}(\boldsymbol{x})}\right)^2 - 1 \leq 0 \tag{28}$$

$$g_4(u_1) = u_1 - u_{1\,max} \leq 0 \tag{29}$$

where ( 29 ) is representative of the maximum tractive force due to engine torque while ( 28 ) includes the maximum tractive force due to braking and bound by the friction ellipse.

Example outputs of the suboptimal ARV final y-position, yaw angle, and yaw rate on the $\mu = 1.0$ surface, can be seen in

Figure 8[1].

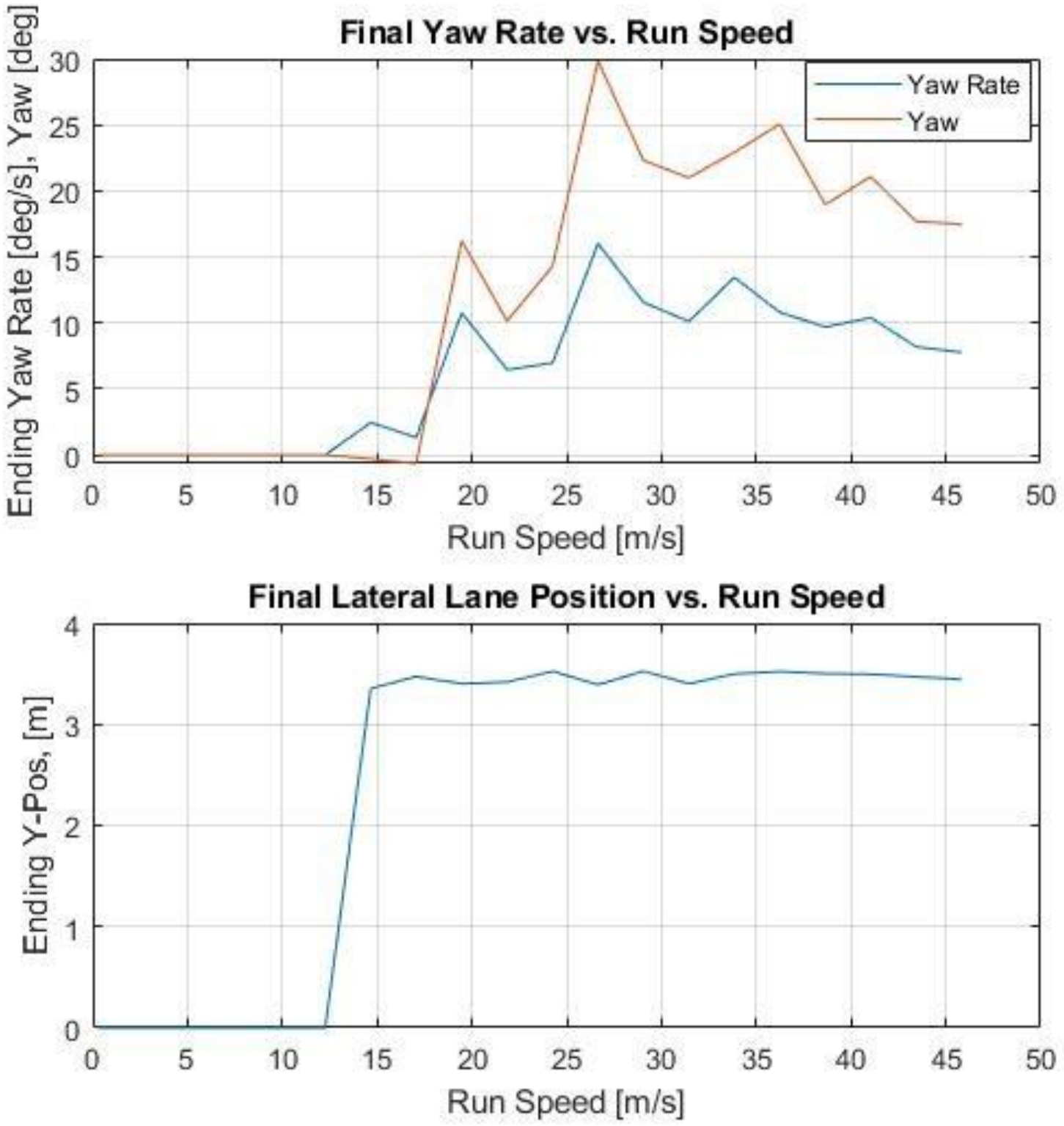

*Figure 8: Example outputs for the EOAM final y-position, yaw, and yaw rate on a high mu surface ($\mu = 1.0$)*

## 4.6 Full Trajectory Profile Creation: Primary Steps

Corresponding suboptimal outputs for the longitudinal acceleration and longitudinal speed profile, for the ARV-ARO relative speed of 100 km/h are shown in Figure 9.

[1] The values in these plots are zero for any speeds below 47 km/h as that was the minimum speed at which an EOAM optimization was performed and considered useful for a highway speed application

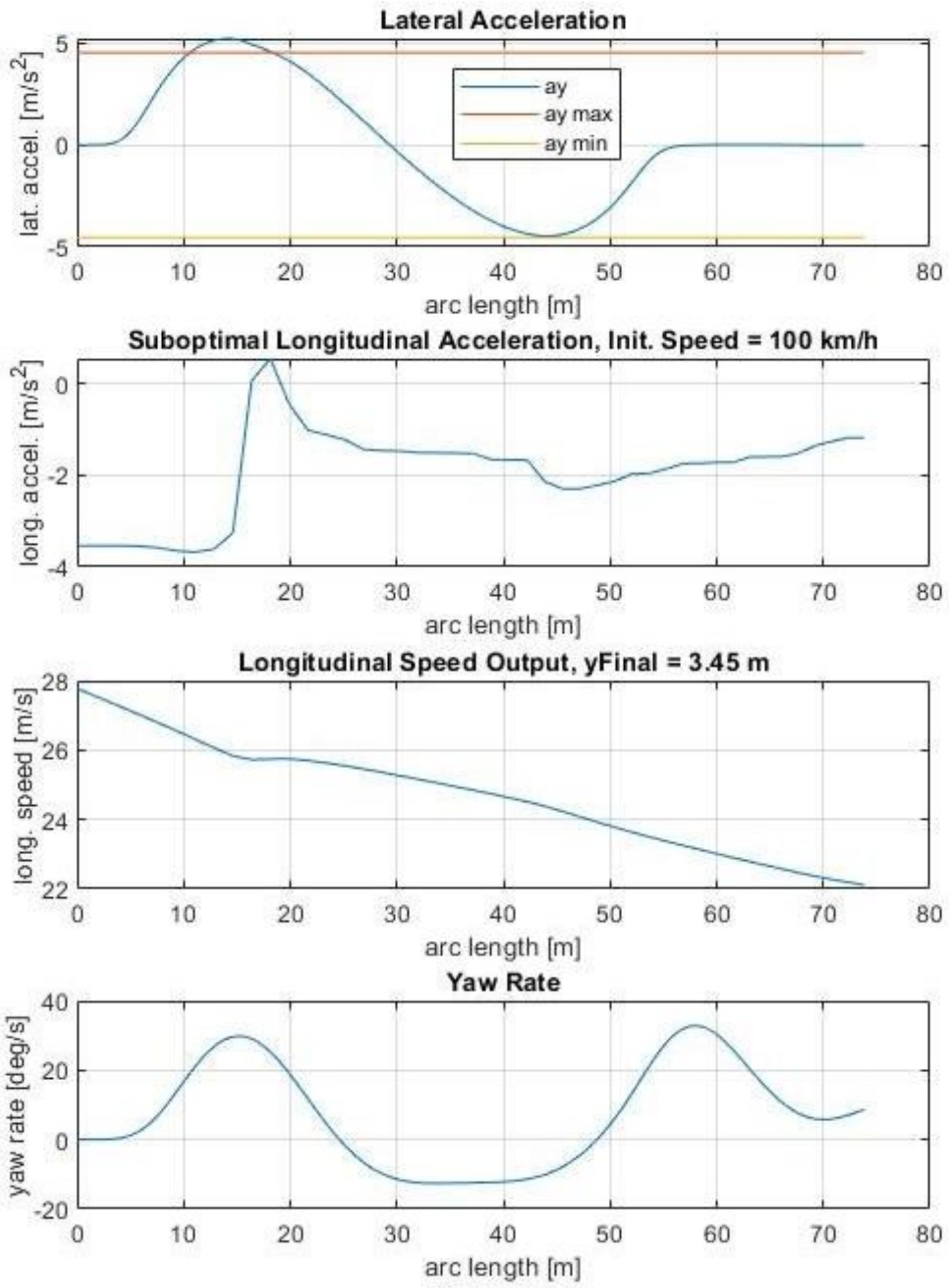

*Figure 9: Sample suboptimal outputs from the longitudinal acceleration optimization at initial speed of 100 km/h*

After the suboptimal longitudinal acceleration profile is determined for a given initial ARV-ARO relative speed, the following steps can be used to complete that DMM phase diagram.

1. Note the final ARV x and y trajectory coordinates, longitudinal speed and acceleration, path curvature, path yaw, steering road wheel angle.
    a. All of the outputs except for longitudinal speed will be used in the 3D lookup tables in the online Simulink model.

b. The longitudinal speed is used only as a reference when checking the output speed control.
2. Calculate the stopping distance for the vehicle based on the initial relative speed.
3. Calculate the minimum clearing distance (the longitudinal distance needed for the ARV front right corner to barely contact the rear left corner of a square object, when completing a left-turn EOAM/lane change) based on the x-y trajectory, path yaw, and expected width of the detected object (outputs and equations shown in next section).
   a. The combined outputs of the minimum clearing distances and the relative speed between the ARV and the ARO at the time of the maneuver, create the EOAM boundary curve in the DMM phase diagram.
   b. The stopping distance at each relative speed is also included in the DMM phase diagram.
4. Repeat the entire process for the next speed in the range of speeds in the EOAM Framework design space, and for the desired tire-road surface friction values meant to be included in the 3D lookup tables.

Once the entire speed range has been simulated with each effective tire-road surface mu, the DMM phase diagrams for each surface mu can be created. The trajectory (paired geometric path and speed profile) outputs for the surface mu of 1.0 are shown below (Figure 10), with more details about the phase diagrams, in subsections to follow.

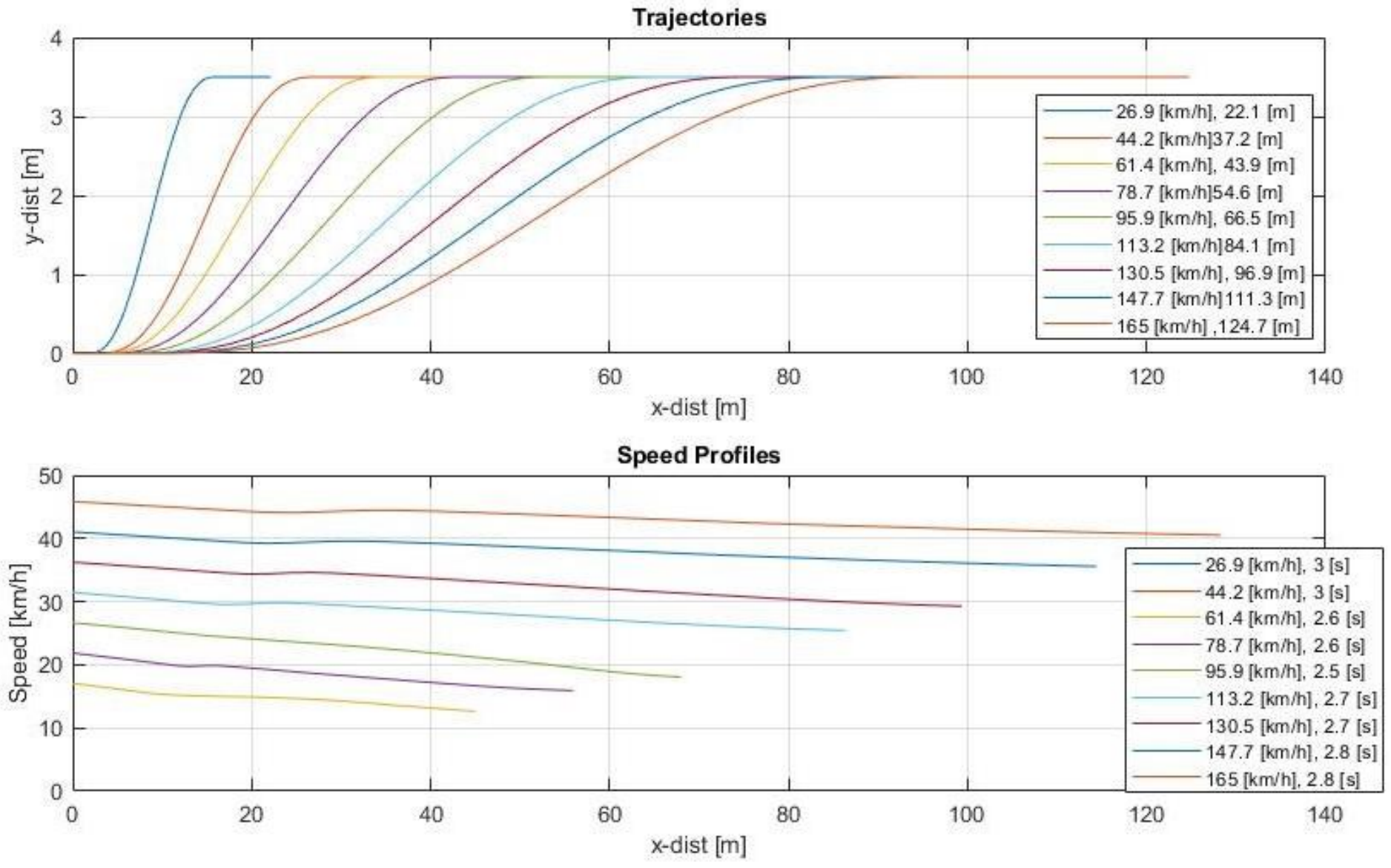

*Figure 10: Suboptimal trajectory and speed profile outputs for a variety of EOAM initial relative speeds, with surface* $\mu = 1.0$

# 5. Decision-Making Module Phase Diagram Creation

The EOAM Framework decision-making phase diagram (DMM) utilizes relative distance and relative speed between the ARV and sensed ARO state variables to determine the EOAM phase, similar to phase diagrams used by (Shiller & Sundar, 1998) and (Hattori, Ono, & Hosoe, 2006). One example (used in the research of this manuscript) of an DMM phase diagram output across a range for relative speeds and on a dry surface ($\mu = 1.0$ can be seen in (

Figure 11).

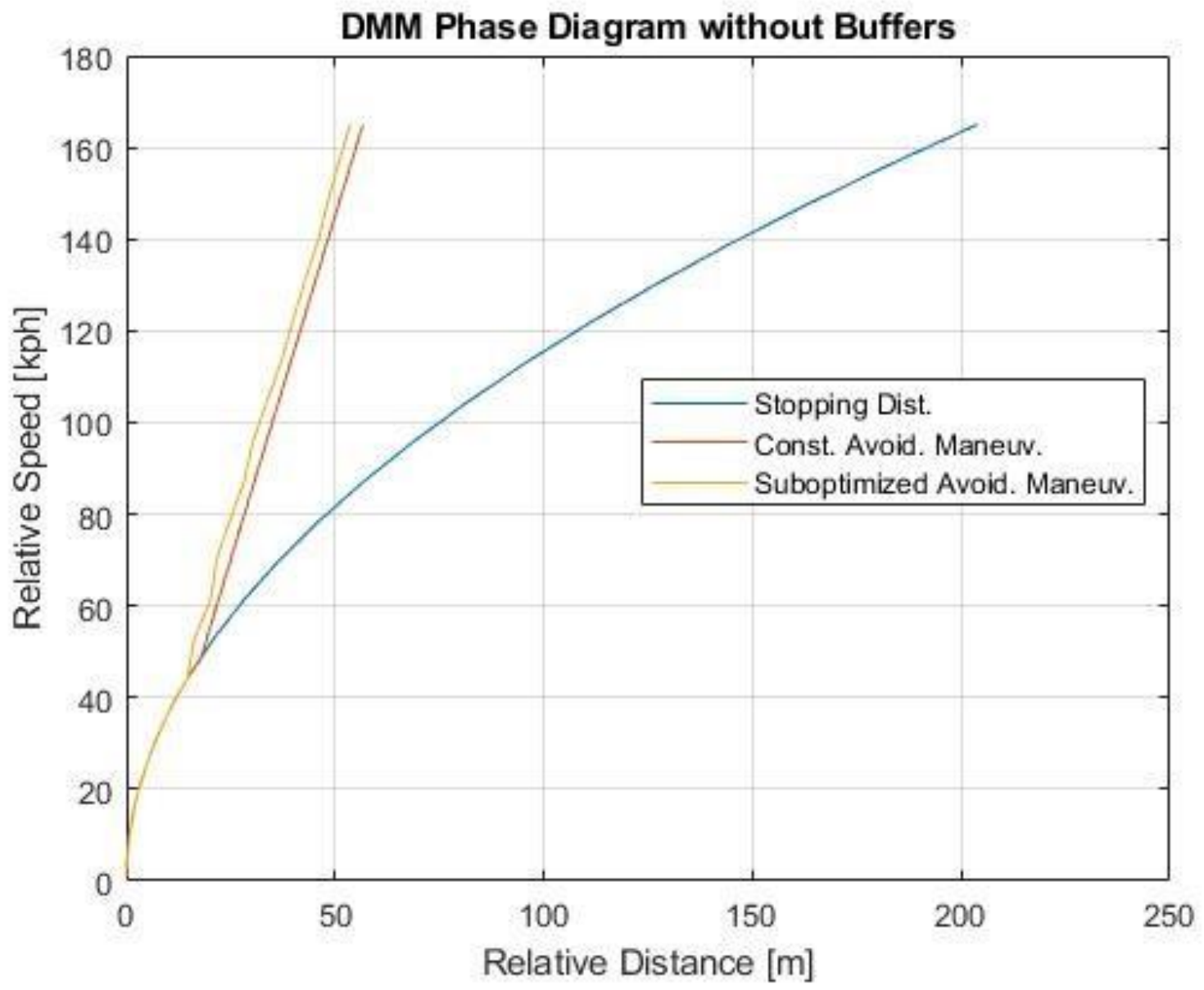

*Figure 11: Initial DMM phase diagram output with stopping distance curve, constant speed minimum clearing distance and suboptimal longitudinal acceleration minimum clearing distance*

This diagram contains phase sections divided by the calculated minimum clearing distances (for both suboptimal longitudinal acceleration and constant speed) and stopping distances at various initial ARV-ARO relative speeds.

This minimum clearing distance is noted as the shortest longitudinal relative distance needed for the ARV to perform a single-lane-change (SLC) EOAM around the ARV, such that during an SLC into a left lane, the ARVs front right corner touches the rear left corner of the ARO at a single point (Shiller & Sundar, 1998). The y-position of the front right corner of the ARV with respect to the rear left corner of the ARO, during the maneuver, can be determined as shown in Figure 12:

$$y_{clearance} = \frac{wid_{obj}}{2} - \left(d_f sin\Psi - \frac{wid_{ego}}{2}\cos\Psi\right) \quad (30)$$

where

$$t_c = t[\min(y_{clearance})] \quad (31)$$

For an example EOAM at 100 km/h on surface with mu = 1.0, this output is follows:

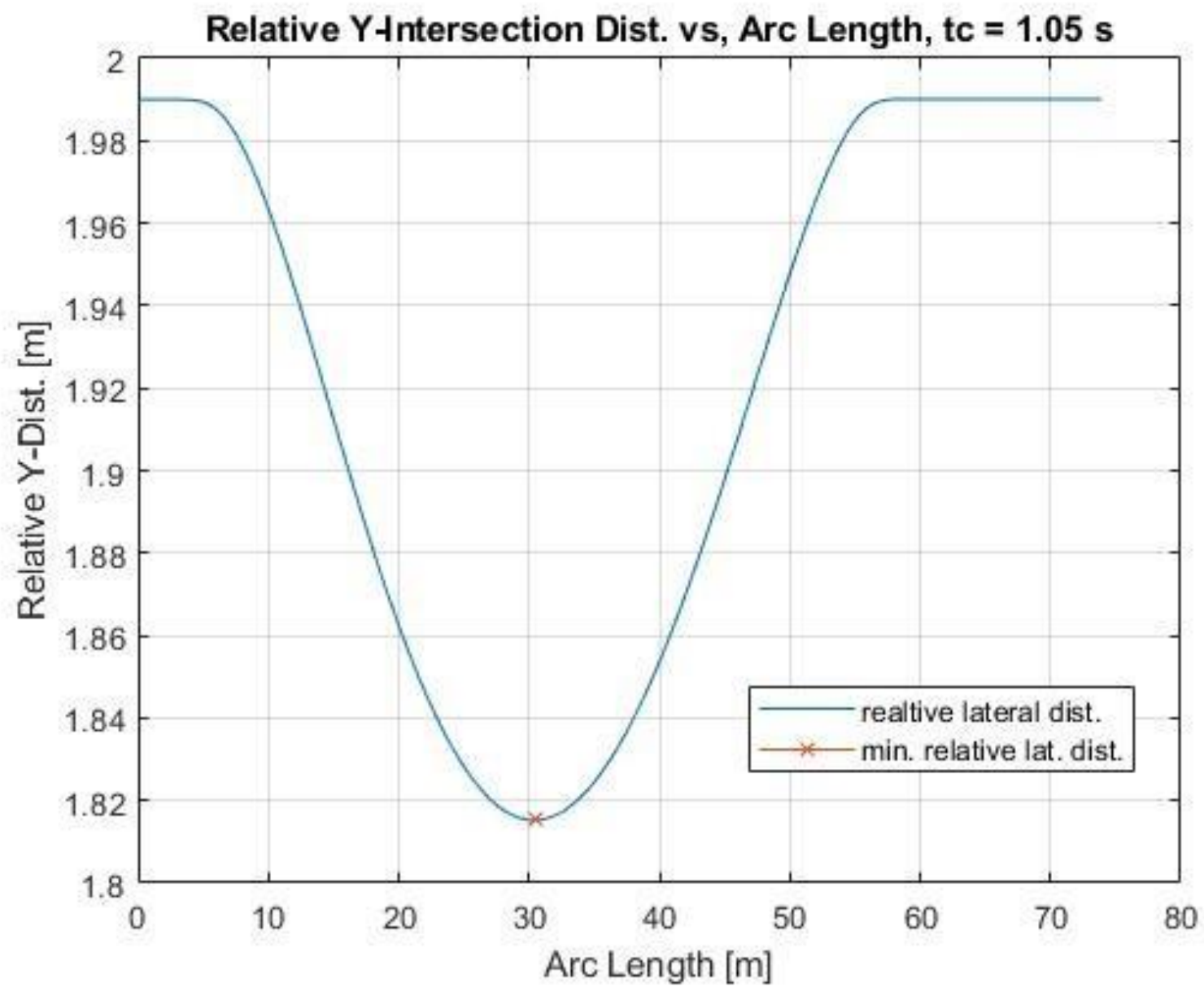

*Figure 12: Relative lateral distance between ARV CG and ARO closest corner, during EOAM*

and the minimum clearing coordinates along the trajectory can be seen on the plot below (Figure 13)

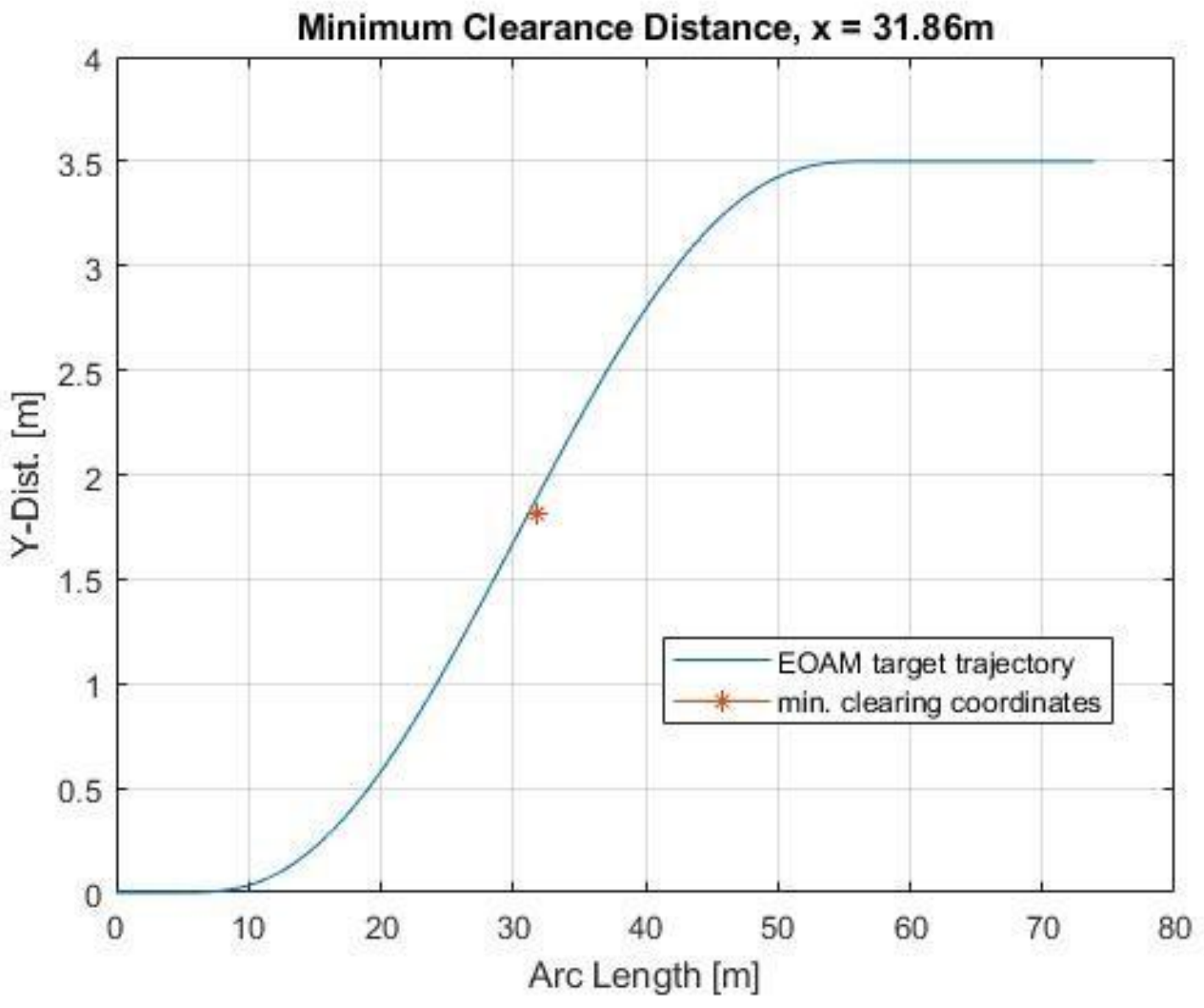

*Figure 13: Minimum clearing distance coordinates superimposed on target CG trajectory*

where

$$x_{clearance} = x(t_c) + d_f \cos\Psi(t_c) + \frac{wid_{ego}}{2} \sin\Psi(t_c) \quad (32)$$

The minimum clearance boundary in the DMM phase diagram is thus created with the minimum clearing distance ($x_{clearance}$) at each initial maneuver speed $\dot{x}_0$ considered for the EOAM ( 32 ). It is also noted that the stopping distance was calculated with the following formulation:

$$x_{stop} = \frac{{\dot{x}_0}^2}{2\ddot{x}_{\max decel}} \quad (33)$$

where $\ddot{x}_{\max decel}$ is based on the vehicle weight and surface mu; this can be determined empirically as the maximum longitudinal deceleration during braking without lockup, as this system is assumed to be without an anti-lock braking system (ABS). The idea that his formulation is based on contact with the ARO — minimal contact, but contact nonetheless — makes the usage of buffers for safety margin clear and necessary. These buffers are used not only to avoid any contact with the ARO during the maneuver but to also reduce the occupant fear and increase their visual comfort of the situation, during the maneuver.

# 6. Novel Decision-Making Module Phase Diagram Updates

The unique DMM phase diagram in this paper embraces the idea that this part of the software logic would be integrated into the system-level ARV architecture as a design domain (Serban, Poll, & Visser, 2018), (Mody, et al., 2018), (Gyllenhammar, et al., 2020). This DMM would be a sub-module of the middle-level EOAM Framework module, as explained in Section 3.

## 6.1 Offline Pre-Computation Approach

The work in this paper calculates the stopping distance and lane change maneuver boundaries offline with the planar bicycle mode, then uses this phase and 3D lookup table data in online implementation (simulation here), to guide a vehicle through the maneuver. This is done by using the boundaries of the phase diagram in the simulated ARV logic so that the ARV recognizes its position in the phase diagram based on its current states and the detected states of the ARO, then makes decisions with unique EOAM logic and the phase boundaries within the decision-making module (DMM) phase diagram.

For a high-level understanding of this type of phase diagram, it is helpful to visualize that the ARV-ARO phase usually moves from right to left, and from high to low, in the DMM phase diagram. This occurs while the ARO is sensed by the ARV’s onboard sensors (radar, camera, lidar, etc.). After the ARO leaves the field-of-view (FOV) of the ARV’s sensors, the DMM phase diagram no longer applies. The ideal final ARV state involves a relative speed between the ARV and ARO to be zero with a relative distance that is greater than or equal to zero, such that a collision is avoided between the ARV and the ARO, during the EOAM.

## 6.2 Novel Phase Boundary Buffers

By including an additional minimum clearance phase boundary for the steering maneuver, phase sectors (B and F) can be utilized that allow the ARV to avoid the ARO ahead, with some buffer distance, and without coming so close to the ARO during the maneuver, as seen in Figure 14.

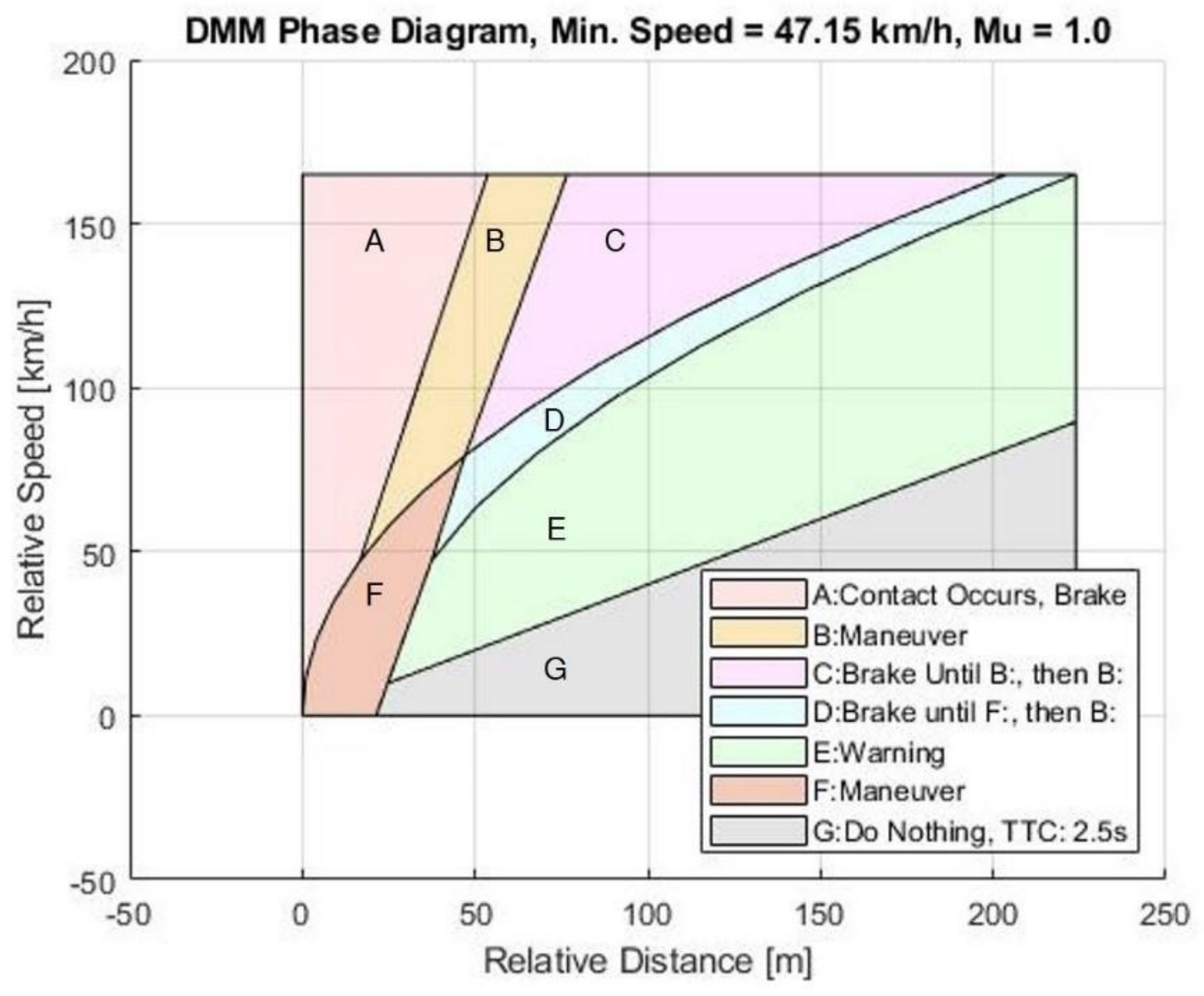

*Figure 14: DMM phase diagram with buffers for braking and steering EOAMs*

This buffer distance is determined by the EOAM Framework system designer and is tunable based on individual performance requirements.

This stopping distance buffer sector formed a braking-only sector in the DMM phase diagram (sectors C and D). If the ARV-ARO phase is within this sector, the ARV applies pure braking until the ARV-ARO phase enters one of the steering sectors. In addition to providing sectors that are dedicated for braking-only maneuver (D), it provides a clear boundary on the phases that require any type of maneuver at all. To the right of this braking-only sector (D), no evasive maneuver is required. However, this braking-only sector is truncated by a maneuver sector represented by the intersection of the steering maneuver and braking maneuvers sectors (F).

The area to the right of the braking-only sector is reserved for applying a visual warning to the vehicle occupants that an emergency maneuver was pending, without performing any evasive action; this sector (E) is to the right of the pure braking sectors. This type of warning is common in ADAS and is sometimes called forward collision warning (FCW). This warning serves to not only give visual and aural indication that a notable object is ahead, but this can also trigger pre-collision systems.

The last sector added was the "do-nothing" sector (G), representing the ARV-ARO phase in which no evasive action is necessary. The boundary for this do-nothing sector is created by the slope of the time-to-collision (TTC) phase boundary.

The TTC phase boundary represents the set of phases in which no evasive action is required by the ARV. The TTC value of 2.5 s was chosen for the high mu (1.0) detected surface condition, where TTC is calculated as follows:

$$TTC = \frac{relative\ distance}{relative\ speed} \tag{34}$$

For the 0.7 mu surface phase diagram, the same TTC was used as the high mu, for 0.3 this was updated to 5 s, and for 0.1 the value was 20 s. These are all tunable values and are left to the discretion of the system designer. It is noted that as the surface mu decreases, the slopes of the lines in the DMM phase diagram based on the representative steering maneuver optimization and stopping distance calculations, also decrease. This decrease in slope equates to the increased necessary distance needed for both steering and braking-only maneuvers to be completed, due to the decreased dynamic longitudinal and lateral force generation capabilities of the ARV (especially due to reduction in tire forces on the road surface).

In summary, the additional relative distance between the ARV and ARO, provided by the buffers in the DMM phase diagram, directly adds safety margin for the ARV and its occupants during an EOAM that utilizes the DMM phase diagram with buffers. As an added benefit, the creation of these buffers also provides additional comfort, indirectly. By potentially triggering maneuvers before the final phase boundary is reached for either stopping or steering and braking, the buffers allow the maneuvers to be completed before/without ARV contacting the ARO ahead. By adding a distance buffer between the point at which the maneuver begins and the minimum clearing distance, the occupants will be less inclined to believe that contact might occur during the maneuver; this is especially true for the steering maneuvers.

# 7. Effective Surface Mu Considerations

It should be noted that values represented as detectable tire-road surface mu can act as an effective tire cornering force and tractive force reduction coefficient, by multiplying the initial linear tire corning stiffness by the mu reduction coefficient. This means that maximum lateral tire force captured by equation ( 6 ) can be adjusted to the following, to reflect the sensed tire-road surface mu

$$F_{y\,max} = -\mu_{surface} C_\alpha \alpha^* \qquad (35)$$

and likewise,

$$F_{t\,min} = \mu_{surface} F_{t\,min-brk} \qquad (36)$$

$$F_{t\,max} = \mu_{surface} F_{t\,max-eng} \qquad (37)$$

for the maximum tractive force due to engine torque and the minimum tractive force due to braking.

This is not an exact representation of the friction between the tire and the road surface, but rather a representation of the possible friction reduction that can occur between a tire and a surface on which it is traveling. There are many approaches to estimating the tire-road friction levels based on available vehicle sensor data, and some of those are listed here for the reader to explore: (Yi, Hedrick,, & Lee, 1999), (Hsu, Laws, Gadda, & Gerdes, 2006), (Nakao, Kawasaki, & Major, 2002), (Svendenius, 2007), (Ahn, 2011), (Güvenc, Aksun-Güvenc, Zhu, & Gelbal, 2021). Outputs for the phase diagram for effective tire-road surface friction can be seen below (Figure 15) for rain on asphalt ($\mu = 0.7$), snow on asphalt ($\mu = 0.3$), and ice on asphalt ($\mu = 0.1$) (Wallman & Åström, 2001).

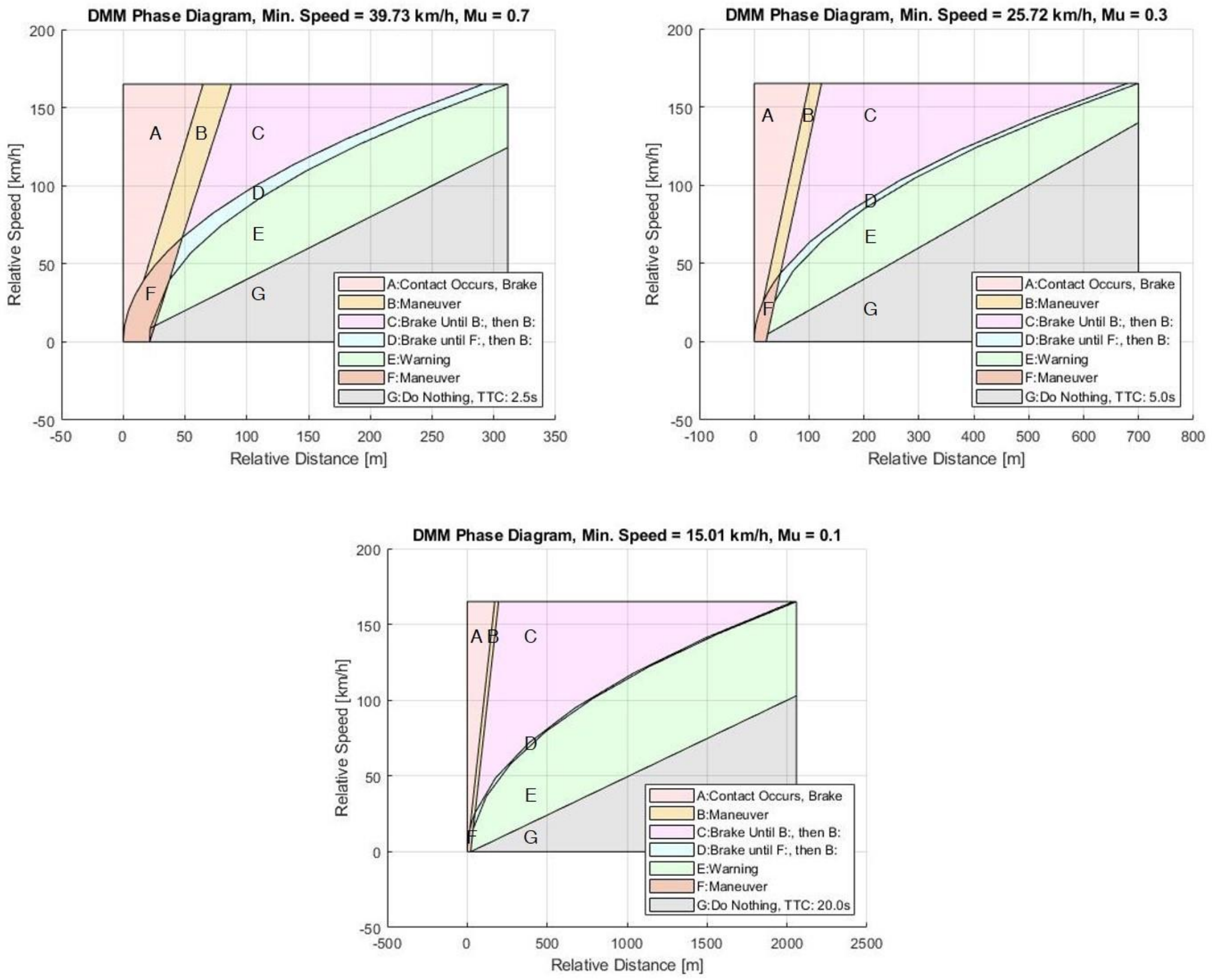

*Figure 15: DMM phase diagram outputs for various effective road surface friction conditions (μ=0.7, 0.3, and 0.1)*

# 8. Other Unique EOAM Framework Features

## 8.1 Point-of-No-Return

During the initial lane-change maneuver, a point-of-no-return lateral threshold value was designated, after which the EOAM logic would complete the lane change maneuver and not allow straight-line braking as an option. Before the point-of-no-return, if the ARV sensed an oncoming vehicle, even if the ARV has initiated a steering maneuver, the sensed oncoming vehicle acts as an override to the maneuver. In essence, an oncoming vehicle sensed by the ARO will trigger emergency braking to avoid making a lane change into oncoming traffic; however, the success of the point-of-no-return critically depends on the *ARV*'s ability to sense and oncoming vehicle or object, with its perception logic.

This point-of-no-return is a lateral threshold from the initial lateral position before the lane change maneuver begins. In the EOAM logic introduced in this paper, the point of no return was calculated as a percentage of the total lateral lane change distance of 3.5 m. The point-of-no-return value was set to 0.3, and 0.3*3.5 = 1.05 m. An illustration of the point-of-no-return ($y_{ponr}$)can be seen in Figure 16.

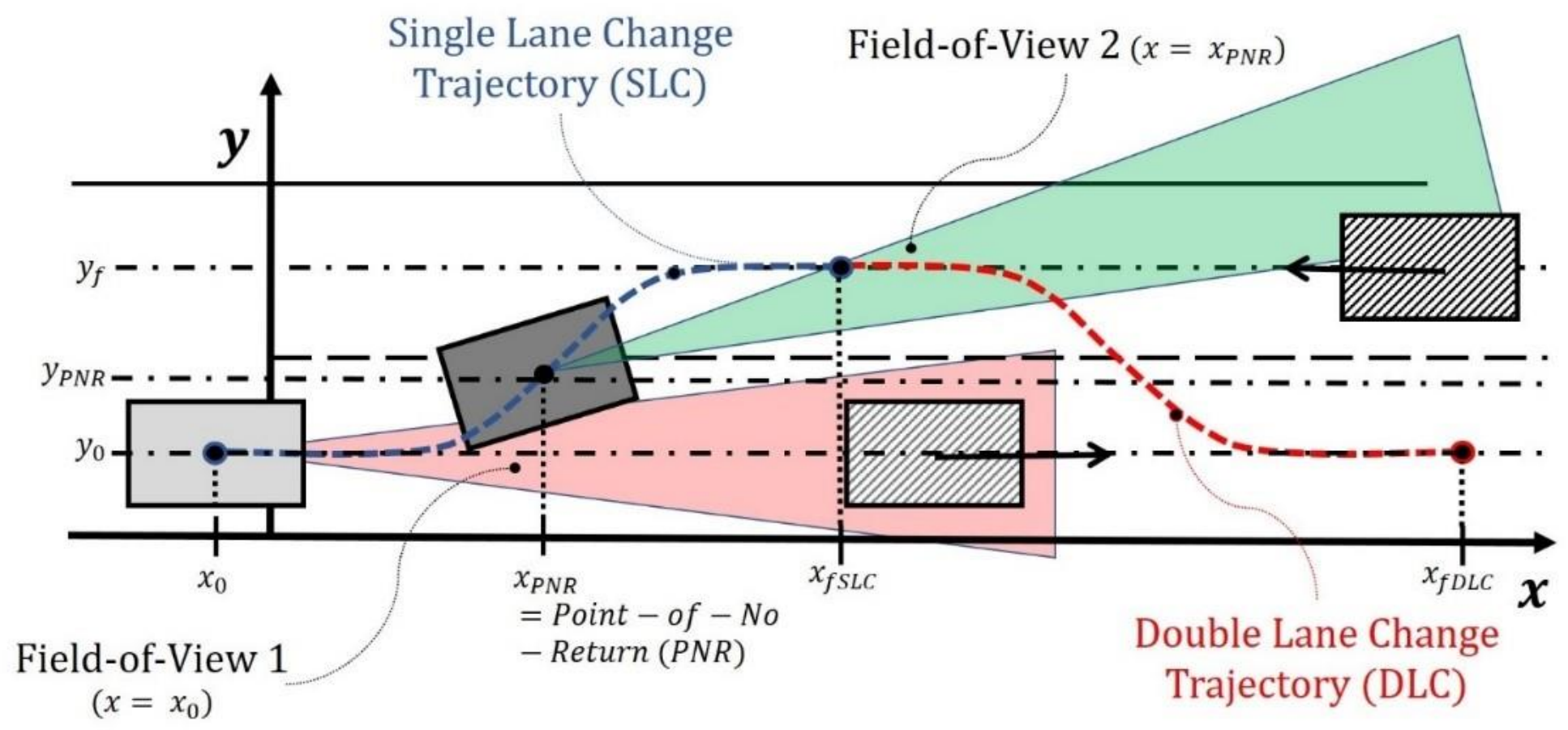

*Figure 16: Top-down view of EOAM with sensor FOV and lateral point-of-no-return boundary*

If the point-of-no-return was utilized, this does not mean that any accident was avoided — on the contrary, it is likely that some contact would occur with the object directly in front of the ARV. However, by choosing to execute straight-line braking instead of a lane change maneuver, the ARV would be utilizing the front crash structures of its vehicle body to better absorb some of the energy in a direct collision compared to the lack of energy offset or oblique collision with an oncoming or adjacent object in the next lane.

## 8.2 Lookup Tables

As mentioned in Section 5, lookup table data are acquired during the process of creating the DMM phase diagram. These data consist of the calculated geometric trajectory for the lane-change maneuver, path yaw angle, path curvature, and longitudinal acceleration, all of which are necessary to calculate the steering maneuver and accompanying optimal or suboptimal speed profile during the maneuver. The steering angle data were not used in a lookup table reference, because the path curvature and control logic were enough to complete the lane change. Alternatively, the steering angle data or instantaneous path curvature and control logic could have been used together, to complete the lane change as well.

Each lookup table contains uniform x-distance and speed values with variable table data — the lookup tables utilized in this EOAM Framework are the following:

- Geometric path for the lane change maneuver
- Longitudinal acceleration profile (coupled with the geometric path to form a trajectory)
- Path yaw angle
- Path curvature

Together, these create two-dimensional lookup tables based on the EOAM initial speed and differential x-distance along the lane change maneuver; the surface mu comprises the third dimension (page data) reference of the lookup tables. For any values that are not explicitly defined in the lookup tables, interpolation is used to determine the appropriate outputs.

An illustration of the 3D lookup table data for the suboptimal longitudinal acceleration profile can be seen in Figure 17: each of the various plot lines represent longitudinal acceleration profiles at various initial relative speed values at the trigger of the EOAM.

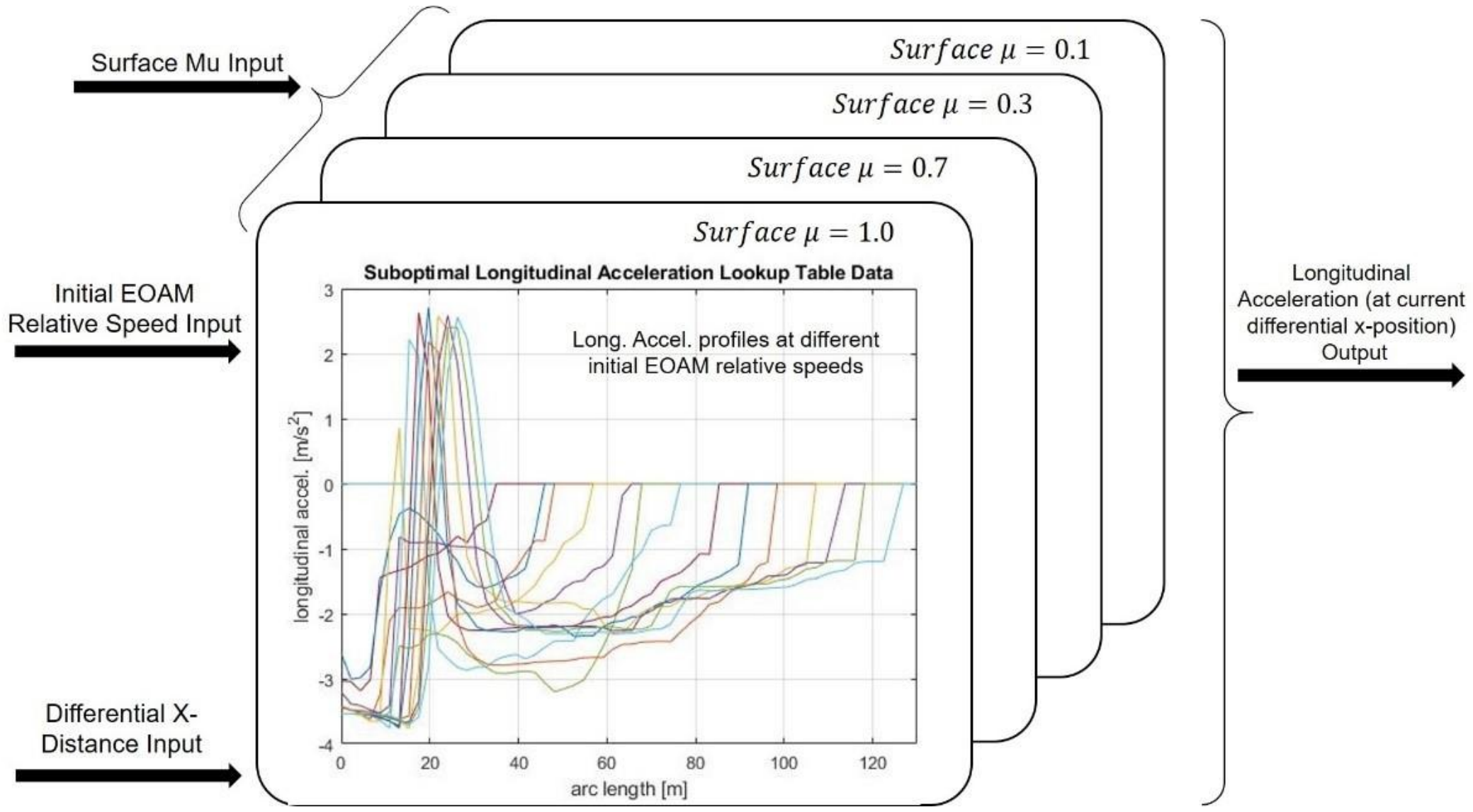

*Figure 17: Illustration of 3D lookup table data for suboptimal longitudinal acceleration profiles at various initial EOAM speeds*

# 9. Control System Architecture

The control systems utilized in this EOAM Framework are applied for the steering and longitudinal acceleration control. The steering controller utilizes three different types of control, in the form of feedforward, feedback, and yaw damping (Lowe, 2022). The feedforward and feedback controllers were previously defined as a potential field lanekeeping controller (Rossetter & Gerdes, 2006), (Talvala & Gerdes, 2008) but can also be used in an application with a defined lane change maneuver as is considered in this paper. Together, the summation of each steering control (feedforward, feedback, and yaw damping) combine to create the overall steering control output:

$$\delta_{control} = \delta_{feedforward} + \delta_{feedback} + \delta_{yaw\ damping} \quad (38)$$

The longitudinal acceleration controller during the EOAM is a PID with longitudinal acceleration error between the reference noted in the longitudinal acceleration lookup table (for the initial

EOAM relative speed) and the actual longitudinal acceleration, at each differential x-position along the EOAM trajectory (Lowe, 2022). While the ARV was not engaged in the EOAM lane change or stopping maneuver, longitudinal speed based PID control was utilized with the initial simulation scenario speed as the reference/setpoint. Note: normally, this speed control would be conducted by the high-level system ARV, and this speed control is representative of the ARV high-level system control, when the EOAM is not active. This PID control of the longitudinal acceleration provided adequate performance with modest controller gains, however, future work would be better suited with applied gain scheduling based on sensed surface mu and vehicle speed.

While a steering controller based on artificial potential fields was used here as it was relatively easy to tune and implement, there are many different control methods that can be adapted and used for steering and also for speed control during the avoidance maneuver. Model Predictive Control (MPC) uses a finite horizon optimal control approach and uses a discrete time model of the plant and can be computed online (Kural & Aksun-Güvenç, 2010) or can be computed offline and stored in tables for real time implementation (Mustafa & Aksun-Güvenç, 2016). The single-track vehicle (bicycle) model used will have uncertainties and changes in transfer function coefficients accompanying changes in speed and cornering stiffness values which require robust control methods. The model regulator also called the disturbance observer can be used to reject yaw moment disturbances while also regulating the plant model about the nominal one (Aksun-Güvenç B. , Güvenç, Ozturk, & Yigit, 2003); (Aksun-Güvenç & Güvenç, 2002a); (Güvenç & Srinivasan, 1994). Parameter space based robust control methods including their frequency domain extension have also been shown to be quite successful in treating these uncertain plants and can also be used together with the model regulator approach (Güvenc, Aksun-Güvenc, Demirel, & Emirler, 2017) ; (Güvenç & Güvenç, 2001); (Aksun-Güvenç & Güvenç, 2002b); (Emirler, Wang, Aksun-Güvenç,

& Güvenç, 2015); (Orun, Necipoglu, Basdogan, & Güvenç, 2009); (Demirel & Güvenç, 2010); (Oncu, et al., 2007); (Güvenc, Aksun-Güvenc, Zhu, & Gelbal, 2021).

# 10. Scenario Creation

## 10.1 Baseline Scenario Setup

The baseline simulation setup in this paper utilized CarSim and Simulink in co-simulation with the DMM phase diagram data and 3D lookup table completed offline in Matlab; the DMM phase diagram boundaries for each surface mu and 3D lookup table parameter matrix data were extracted from the Matlab Workspace and utilized in the Simulink DMM Main Logic block, for online simulation.

Individual ARV speeds of 165, 120, 90, and 55 km/h were used in the baseline simulation test matrix, along with surface friction values of 1.0, 0.7, 0.3, and 0.1. The layout for the baseline runs were conducted on a simulated straight two-lane highway in a proving ground setting (CarSim 2021.0), where the adjacent lane allows for traffic traveling in the opposite direction.

In addition to the road surface, parked cars were added to the right of the highway lane on which the ARV was traveling (1.5 m from the right lane marker) as shown in Figure 18, to assure that the ARV (with radar sensor) could perform the EOAM without detecting parked cars as false-positives. For each baseline scenario, the ARV encountered the same ARO, D-Class Minivan, which started at 120 m ahead of the ARV and at an initial speed of 60 km/h (Figure 18).

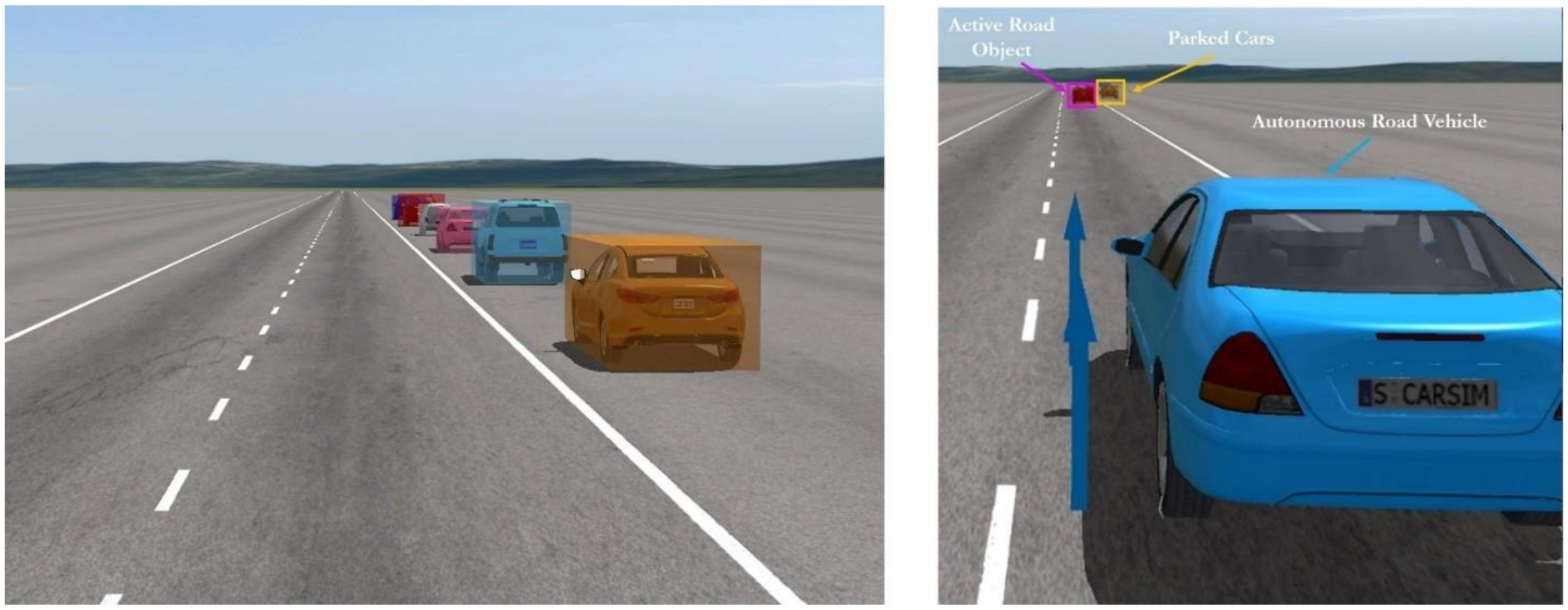

*Figure 18: ARV shown with ARO (outside of ARV radar range) in its lane of travel, with parked cars to the right side of the marked lanes*

This ARO traveled on a straight path in the same lane as the ARV, and executed the following speed profile, to mimic an emergency stop on the highway (Figure 19):

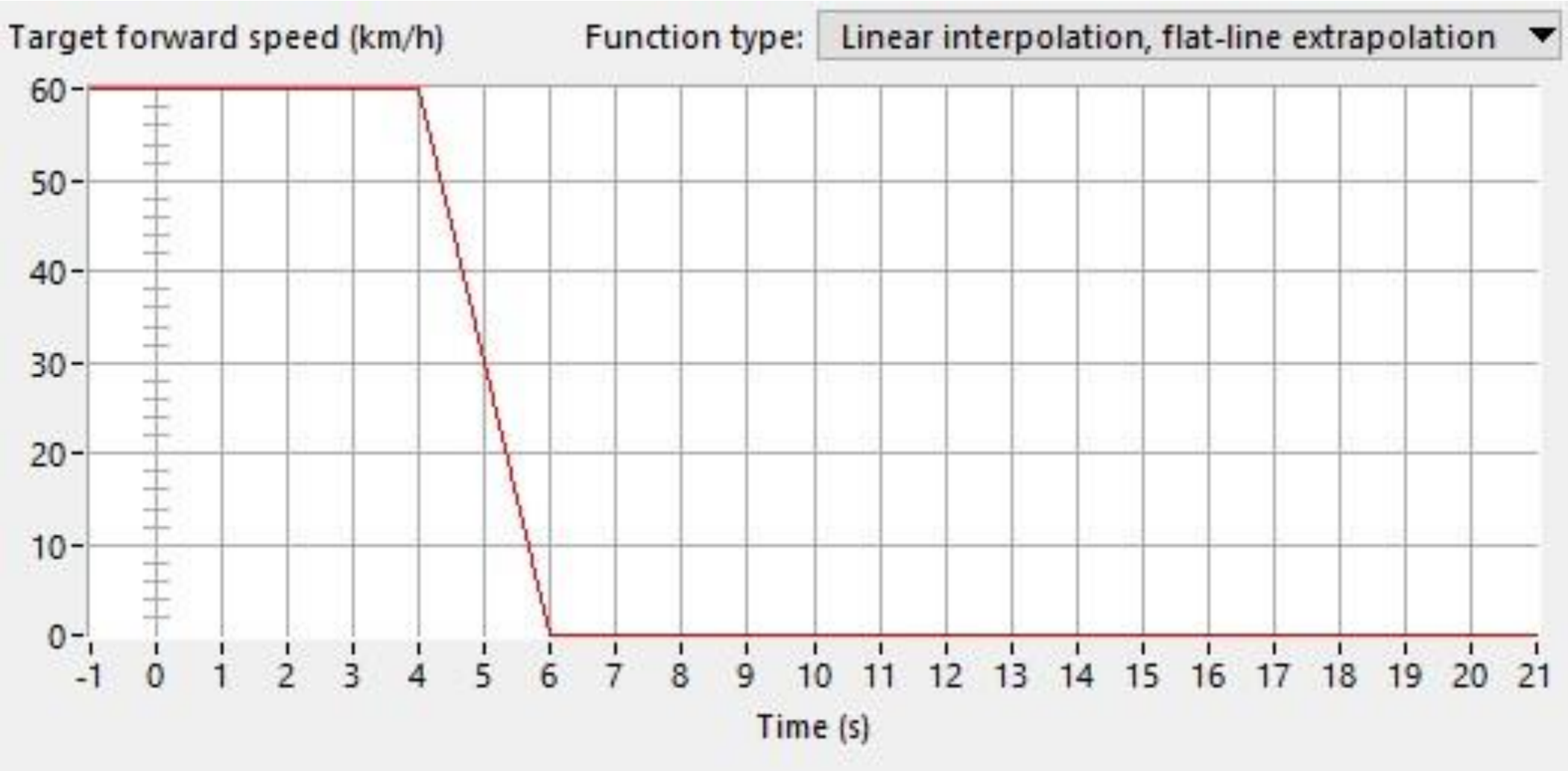

*Figure 19: Longitudinal speed profile for ARO (D-Class Minivan) in front of the ARV in each test scenario*

The combination of proving ground setting with two-lane highway, parked cars for sensing and perception ARO false-positive check, and ARO which conducts an emergency stop in the same lane as the ARV, comprised the fundamental pieces of the baseline test scenarios. Depending on the initial speed of the ARV, the execution of the EOAM by the ARV followed the same steps as listed in the EOAM Framework Logic section. Below are stroboscopic views of the ARV performing EOAM at 165 km/h with dry surface ($\mu = 1.0$) (Figure 20).

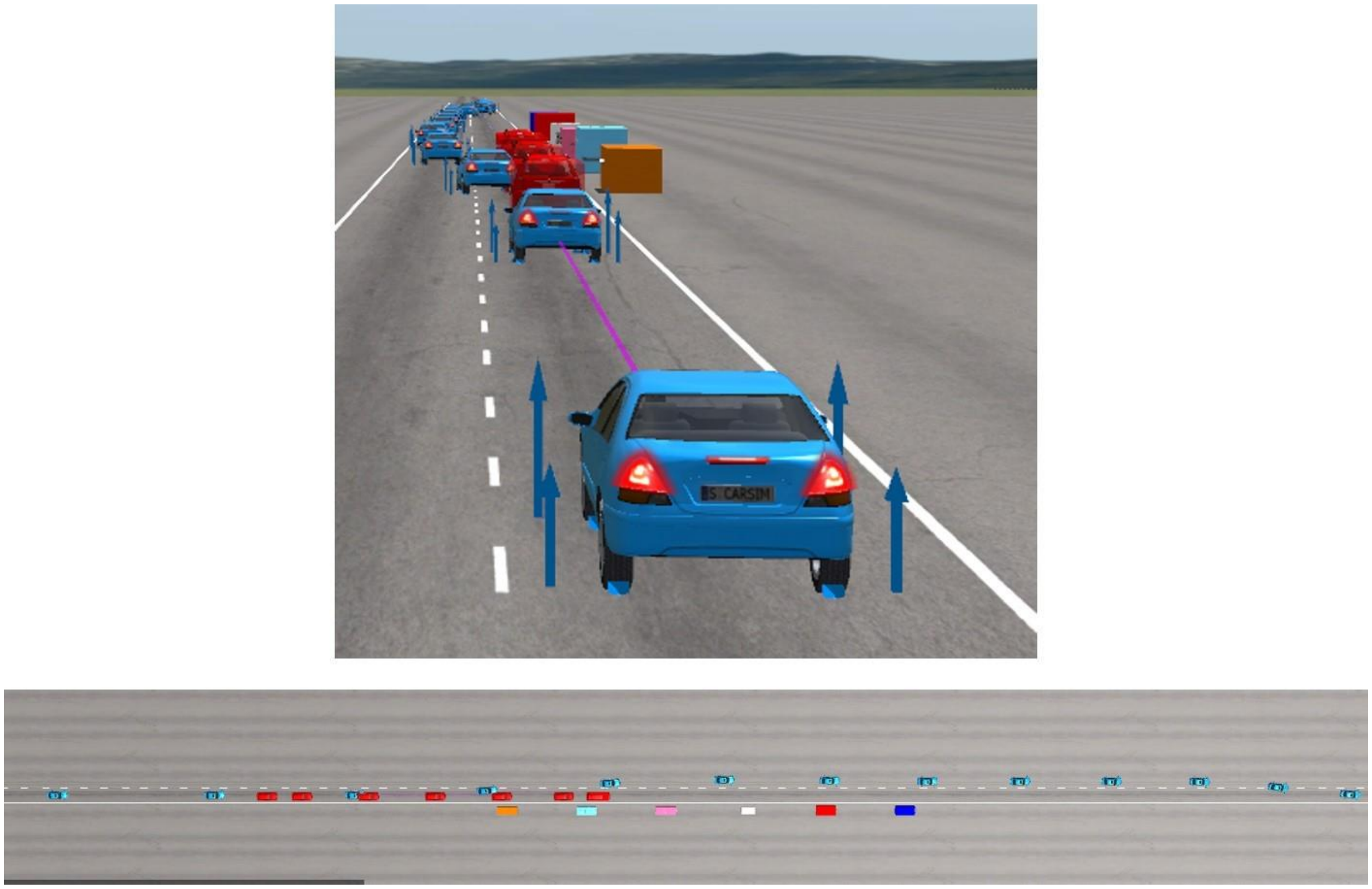

*Figure 20: Rear and top-down stroboscopic views of the ARV performing an EOAM at 165 km/h on the dry surface*

## 10.2 Oncoming Vehicle Scenarios

In addition to the baseline scenarios listed above, the same scenarios were attempted with an oncoming additional ARO traveling in the lane adjacent to that of the ARV in its initial position. The scenarios with the oncoming ARO were repeated with the oncoming ARO starting at 500 m, 400 m, and 300m away from the ARV. In each test scenario including the oncoming ARO, the ARO speed was constant at 20 m/s (76 km/h). Each oncoming vehicle scenario was tested with the discrete ARV test speeds of 165, 120, 90, and 55 km/h, and discrete surface mu conditions of 1.0, 0.7, 0.3, and 0.1.

# 11. Results and Discussion

The results summary of the test scenarios are shown below in Table 2, with the following color-coding definitions:

- **Green**: the EOAM single lane change (SLC) was performed successfully without collision of any type, and return to the original lane.
- **Yellow**: the EOAM logic worked as planned but an imminent collision was detected based on the DMM phase diagram so straight-line limit braking was applied and a collision occurred with the ARO ahead, OR

  the ARV detected oncoming traffic before the point-of-no-return and determined that straight-line limit braking was necessary to avoid a collision with oncoming traffic, and by doing this had a reduced-speed direct (rather than offset/oblique) collision with the ARO.
- **Orange**: the ARV performs the lane-change EOAM and there is some resulting lateral contact with the oncoming vehicle (not a head-on collision).
- **Red**: the ARV performs the lane-change EOAM resulting in a head-on collision with the oncoming vehicle that is either direct (squarely in front of the offset vehicle), or offset/obliquely.

## 11.1 Dynamic Performance Output for 120 km/h ARV EOAM Baseline Scenarios

A sample of the ARV dynamic performance can be seen when viewing the output data for the 120 km/h EOAM scenario on the dry surface ($\mu = 1.0$), without oncoming traffic DMM phase diagram output (Figure 21).

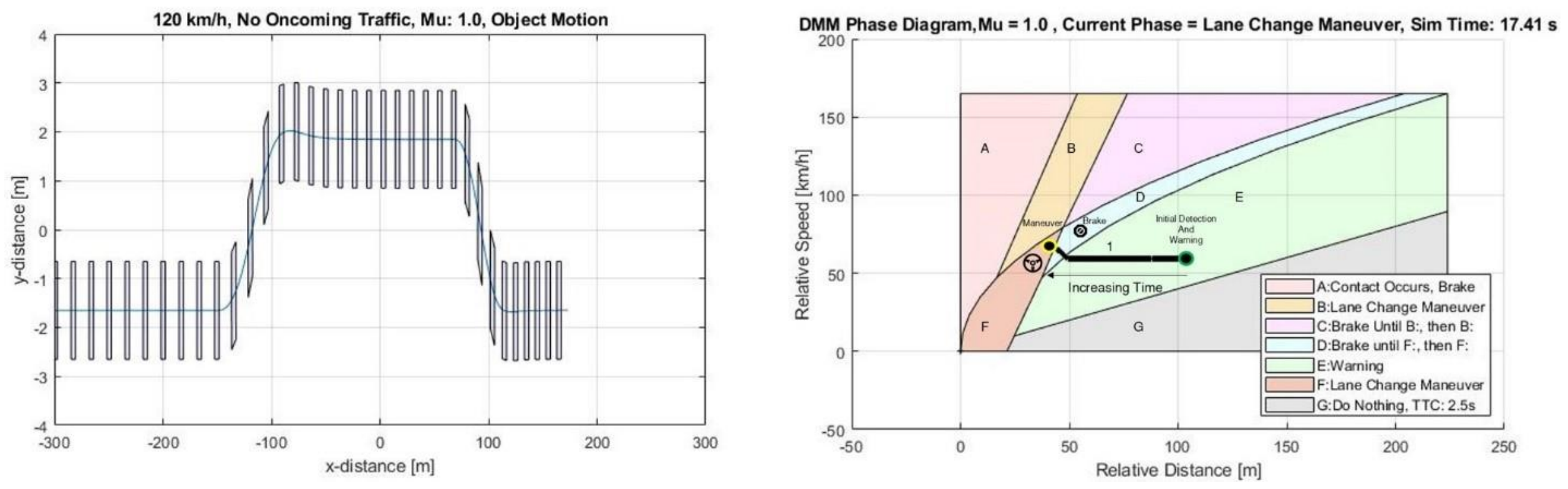

*Figure 21: Motion of the ARV and ARO, and DMM phase diagram for dry surface and 120 km/h scenario and no oncoming traffic, with phase data labeled as trace 1*

Outputs in Figure 21 demonstrate that the EOAM performed by the ARV allowed it to successfully avoid the ARO, and without the EOAM, a collision would have occurred. In Figure 21, the black phase data (trace 1) indicates that upon first sensing the ARO, the ARV is in the FCW phase (sector E) of the DMM phase diagram to warn occupant(s) of an impending EOAM and arm any available pre-collision safety systems. When appropriate the ARV reduces its speed by braking while in sector D, to reduce the severity of the upcoming steering maneuver, then performs the steering maneuver upon entry into sector F of the DMM phase diagram. It is noted that choosing a steering maneuver in sector F allows the ARV to clear the ARO with more distance (and associated passenger visual comfort) than if a pure braking maneuver were performed instead. The phase trace truncates in sector F when the FOV of the ARV radar sensor no longer detects the ARO. There are several other dynamic output plots describing the EOAM modes-of-operation, forward collision warning (FCW) flag, longitudinal speed and acceleration profiles versus the target values, the ARV EOAM actual versus target lateral position, lateral offset and lookahead error outputs, and differential x-position output during each steering maneuver and used for the 3D lookup tables (Figure 22).

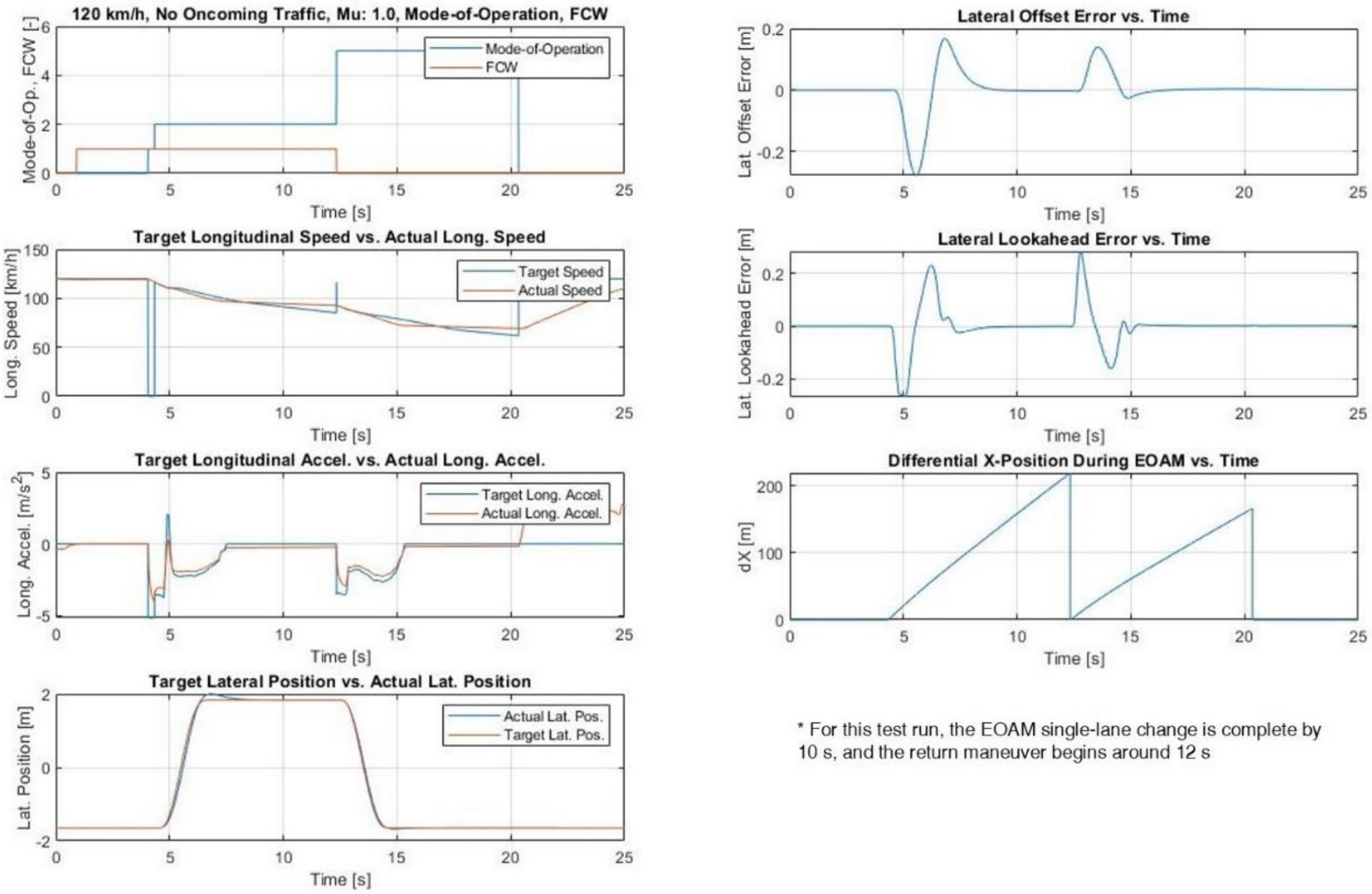

*Figure 22: Various dynamic outputs from the 120 km/h EOAM scenario on a dry surface without oncoming traffic: EOAM mode-of-operation, FCW flag, target vs. actual longitudinal speed, target vs. actual longitudinal acceleration, target vs. actual lateral position, lateral offset error, lateral lookahead error, and differential x-position during the EOAM (for the 3D lookup tables)*

The modes-of-operation shown in Figure 22 indicate that, during the EOAM, first a braking maneuver was applied (1), followed by the steering maneuver with longitudinal acceleration control (2). After the single lane change (SLC) steering was completed and when the maneuver timer reached its maximum time, the ARV completed the return maneuver to the original lane (5). At the speed of 120 km/h the lateral position control was adequate, with only slight overshoot indicated upon reaching the desired lateral position for the SLC; that amount of overshoot is expected for a maneuver at this speed. The lateral offset and lookahead error are also within expected bounds. The differential x-position output is offered just to show the values that are fed into the 3D lookup tables during the EOAM steering maneuver, to extract the various outputs (target lateral position, longitudinal acceleration, path yaw, path curvature) utilized for the ARV control. This differential x-position resets when the steering maneuver is fully complete (8 s duration) or when the return maneuver begins. Outputs including the steering controller roadwheel angle, combined steering handwheel angle (steering ratio is applied), corresponding lateral acceleration and yaw rate are seen in Figure 23.

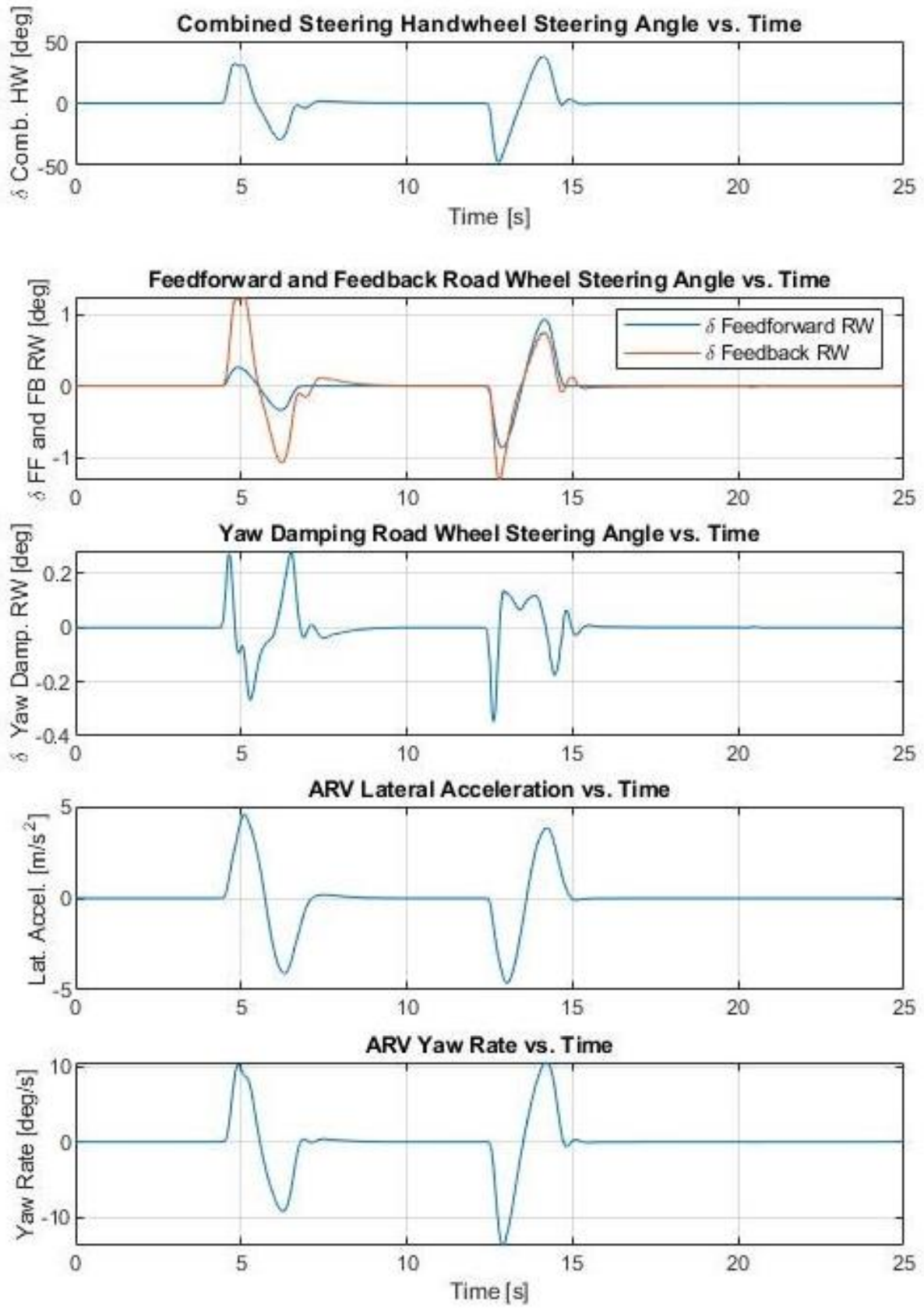

*Figure 23: Steering, lateral acceleration, and yaw rate outputs for the 120 km/h ARV EOAM scenario on a dry surface without oncoming traffic*

From the steering outputs, it is apparent that the controllers work well together to allow the ARV to follow its desired trajectory during the ARV. The feedforward control for the initial EOAM steering maneuver is small in magnitude and this is due to the steering maneuver optimization which does not account for steering control to be applied; instead, the optimization is expecting only open loop steering application which would result in more lateral motion than what is seen with the closed-loop steering control. Thus, the feedback steering control for this maneuver increases the steering angle to assist the ARV in reaching its final desired lateral lane position for the SLC.

The yaw damping controller is quite active for the initial maneuver due to lateral acceleration and yaw rate outputs at the traction limit of the tire. The yaw damping steering controller also provides useful steering input for the RETURN maneuver, even though the ARV longitudinal speeds during the RETURN are lower than those during the initial SLC. There is an indication of the ARV's longitudinal versus lateral acceleration output for the 120 km/h dry surface scenario with no oncoming traffic as can be seen with the E-class sedan's parameter-based friction ellipse superimposed in (Figure 24); sometimes this is called a g-g diagram (Milliken & Milliken, 1995)

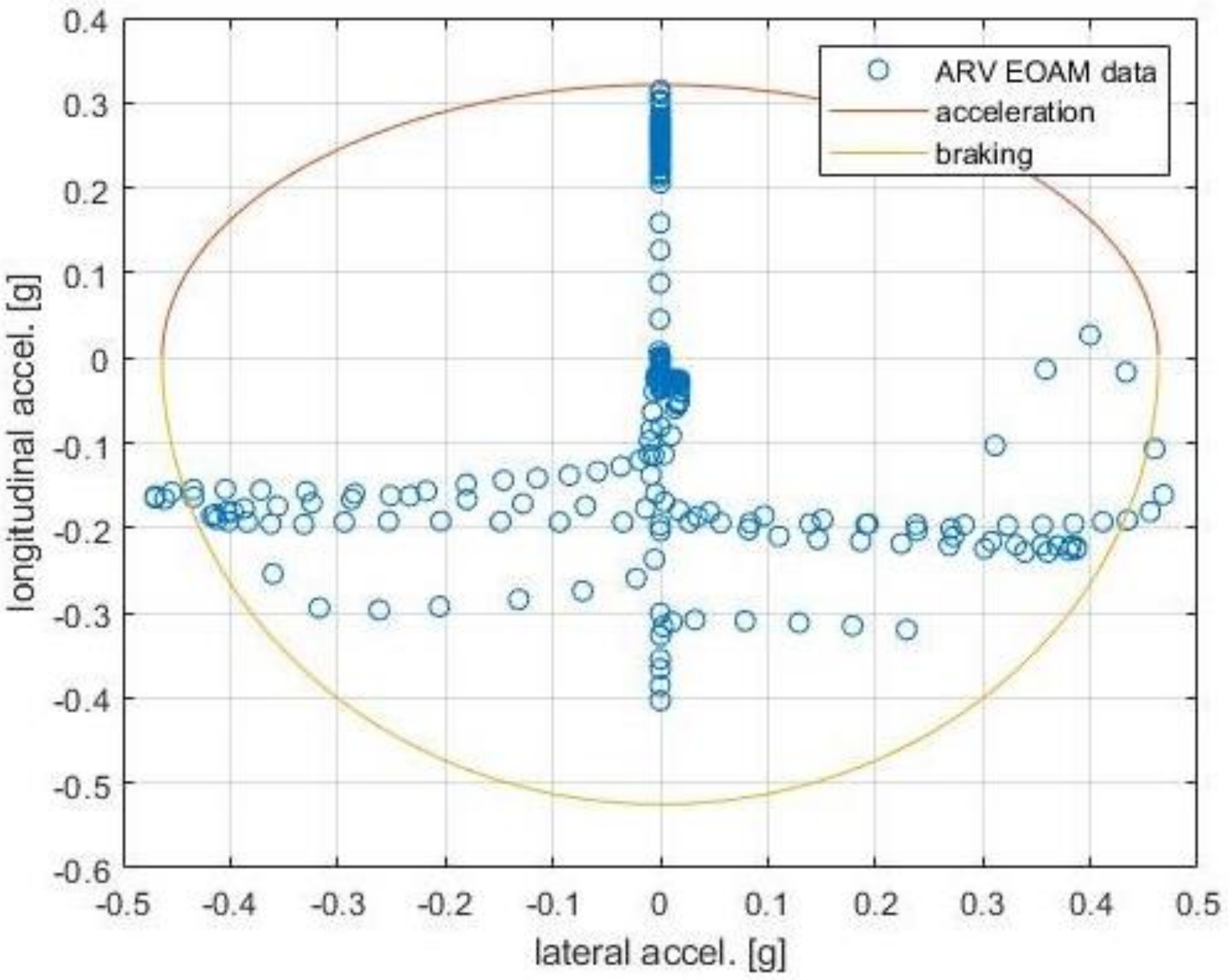

*Figure 24: Longitudinal vs. lateral acceleration ARV outputs with the E-Class sedan parameter-based friction ellipse superimposed, for the 120 km/h ARV EOAM scenario on a dry surface without oncoming traffic*

The lateral and longitudinal acceleration outputs for this 120 km/h dry surface scenario without an oncoming vehicle mostly lie within the friction ellipse, though some of the lateral acceleration outputs lie just outside of the ellipse during the steering and RETURN portions of the EOAM; these excess lateral acceleration magnitudes (less than 0.03 g) can be attributed to the fact that the trajectory output is suboptimal in nature. This was also expected based on suboptimal lateral acceleration results in the creation of the DMM phase diagram, as was shown in Figure 9.

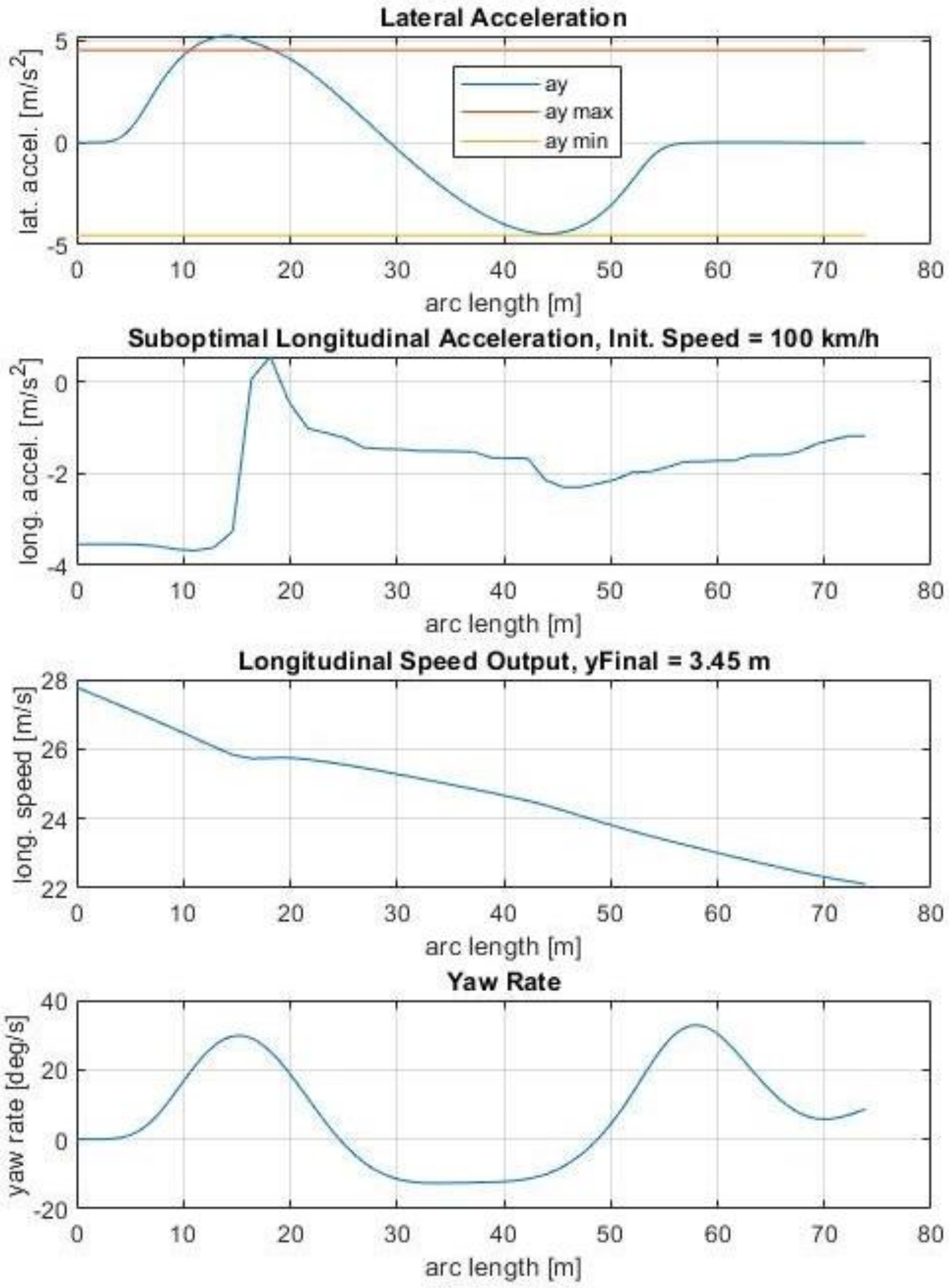

Figure 9.

To understand a snapshot of broader ARV EOAM performance, outputs of the ARV trajectories during the EOAMs are shown for the ARV traveling at 120 km/h on all of the various surface mu values (1.0 through 0.1) are shown in the left plot of Figure 25 and outputs on the dry surface with various oncoming traffic scenarios are shown in the right plot.

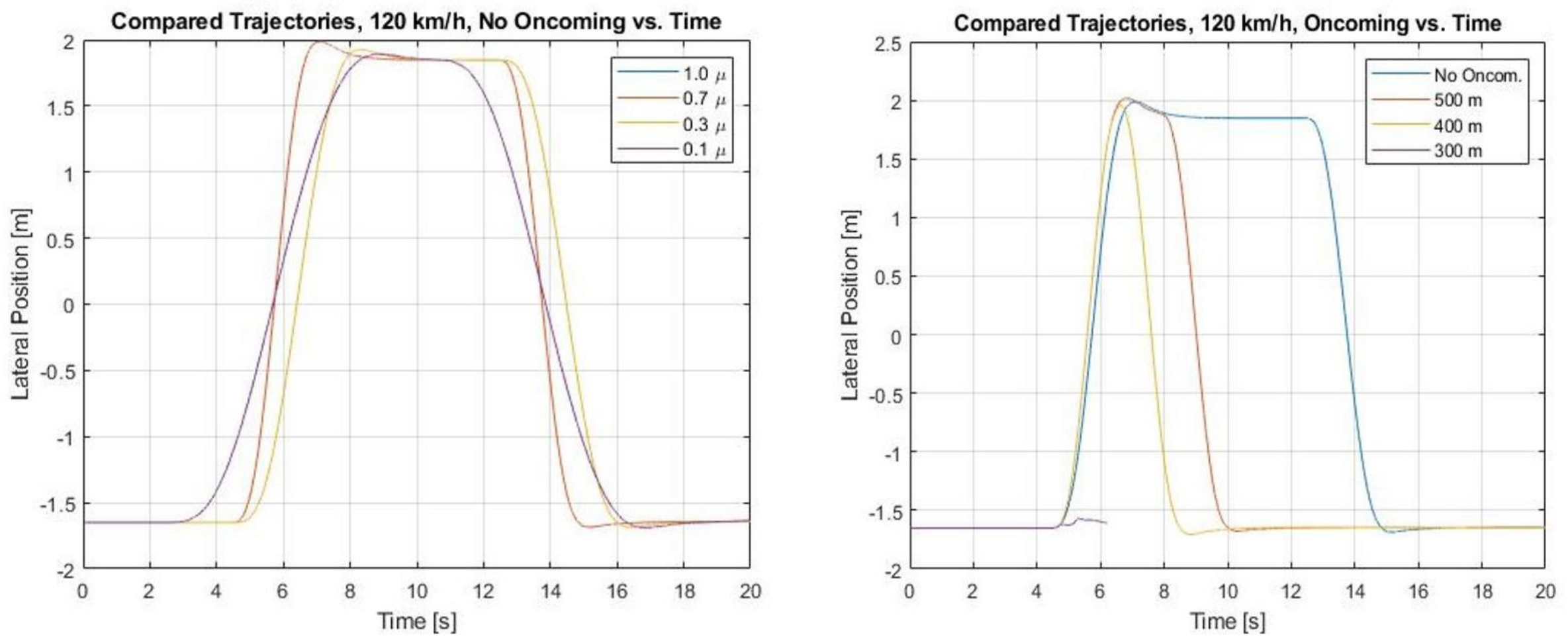

*Figure 25: (Left) Output ARV lateral position profiles vs. time for 120 km/h EOAM scenarios on various surface friction values without oncoming traffic and (right) ARV lateral position during 120 km/h EOAM with oncoming traffic starting at varying distances from the ARV's initial position (500 m, 400 m, and 300 m away) on the dry surface*

The trajectory profiles in Figure 25 show that 1.0 and 0.7 surface mu values yield nearly identical and not distinguishable results in this plot. It is reasonable to conclude that the EOAM SLC outputs for a high mu dry surface (1.0) and high mu wet surface (0.7) would yield similar EOAM SLC outputs in EOAM framework as designed in this paper.

The outputs in Figure 25 should be compared to the 120 km/h experiment summary in Table 2. What is notable is that the addition of the oncoming vehicle causes the SLC to be truncated before the maximum maneuver time; this is due to the RETURN being triggered early upon perception of the oncoming ARO, by the ARV. This is the expected output based on the proscribed EOAM Framework logic. Additionally, the test scenario with the oncoming ARO starting only 300 m from the ARV indicates graphically that the ARV sensed the oncoming ARO before the point-of-no-return (twice); then, the ARV performed limit braking to minimize the danger of the EOAM by choosing a collision with reduced speed that best uses the ARV's crash structures and safety devices (airbags, etc.).

## 11.2 Full EOAM Simulation Results Summary

The overall output for the runs in the full ARV EOAM experiment can be seen in Table 2. This table uses the same color distinctions mentioned at the beginning of this section.

*Table 2: ARV EOAM All Experiments Results Summary*

| Traffic | Initial ARV Speed | 1.0 surface mu | 0.7 surface mu | 0.3 surface mu | 0.1 surface mu |
|---|---|---|---|---|---|
| No Oncoming | 165 km/h | green | green | green | yellow |
| | 120 km/h | green | green | green | green |
| | 90 km/h | green | green | green | orange |
| | 55 km/h | green | green | green | green |
| Oncoming, 500 m | 165 km/h | green | green | red | yellow |
| | 120 km/h | green | green | orange | yellow |
| | 90 km/h | green | green | green | orange |
| | 55 km/h | orange | orange | yellow | red |
| Oncoming, 400 m | 165 km/h | green | green | red | yellow |
| | 120 km/h | green | orange | orange | orange |
| | 90 km/h | orange | orange | yellow | yellow |
| | 55 km/h | orange | green | green | yellow |
| Oncoming, 300 m | 165 km/h | yellow | red | red | yellow |
| | 120 km/h | yellow | yellow | yellow | yellow |
| | 90 km/h | green | green | green | orange |
| | 55 km/h | green | green | green | green |

| | |
|---|---|
| red | EOAM performed, oncoming collision occurred |
| orange | EOAM performed, lateral collision occurred |
| yellow | Limit braking applied with expected front collision |
| green | Successful EOAM SLC with return to lane |

What's most notable about the full experiment output table is that the majority of the 0.3 and 0.1 surface mu values involve collisions, either those that were anticipated and prepared for with straight-line limit braking before the point-of-no-return or more severe collisions that occurred with the oncoming ARO. This is one area where the experiments in this manuscript extend beyond other studies that have a single ARO in the lane of the ARV and no oncoming traffic. This summary highlights the importance of considering oncoming vehicles for an EOAM when deciding if an EOAM should be executed and if so, what type of maneuver.

One positive note regarding the collisions that occurred in the scenarios with oncoming vehicles is that most of them involved avoiding the oncoming vehicle by utilizing the point-of-no-return feature. This shows that the addition of the point-of-no-return feature added an extra element of safety for the ARV and its occupants than if it were not included. The other collisions that involved the oncoming ARO are important to highlight, as these could have catastrophic outcomes (critical injuries and fatalities) for the ARV and its occupants. These collisions, however, could be eliminated with an additional vehicle-to-vehicle (V2V) or vehicle-to-infrastructure (V2I) communication, notifying the ARV that oncoming traffic exists and triggering the same in-lane, straight-line limit braking that can be triggered by the point-of-no-return feature.

# 12. Conclusions and Recommendations

With the details presented in this manuscript, it is apparent that the proposed EOAM Framework for ARVs is useful and effective in providing a domain-based emergency maneuver system for ARVs designed to be integrated into other functional ARV system architectures. This EOAM framework was shown to function within a high-level ARV system that required taking control of the ARV during the EOAM, then handing control back to the high-level system, once the EOAM was completed. Additionally, the inclusion of a point-of-no-return feature that depended on the sensing and perception capability of the ARV to detect oncoming traffic, was a useful and necessary EOAM feature to minimize contact with an ARO if it was imminent. The various features of the EOAM framework and accompanying logic included offline-calculated data used to create the proposed DMM phase diagrams and 3D lookup tables which were crucial in allowing the ARV EOAM Framework to conduct sensing, perception, decision-making, trajectory generation, control, and actuation, quickly enough to avoid the ARO ahead. This capability for integration into larger ARV system architectures for daily road usage makes this EOAM

Framework novel as compared to other proposed frameworks that do not focus on taking control and handing it back to other system domains within the ARV's architecture.

While the simulation experimentation for the ARV EOAM Framework highlighted in this manuscript was extensive with only some exemplary results being presented, there are other opportunities for future work in this research area, including but not limited to the following:

1) Creating EOAM Framework lookup table data for a vehicle including ABS and ESC, then seeing how it performs in the tests shown in Table 2.
2) Perform EOAM scenario tests with other traffic scenarios on other highway types with various levels of traffic
3) Inclusion of ARV EOAM Framework logic updates that utilize V2X.
4) Inclusion of actual sensing and perception systems.
5) Conducting experimentation with different types of vehicles (compact cars, trucks, vans, SUVs) with different drivetrain layouts (FWD, AWD, electronic-AWD).

All of these items would illuminate a rich area of research relating to ARV EOAM capabilities and it is recommended that these tests be explored before mass utilization of ARVs on public roads at highway speeds.